\documentclass[11pt,letterpaper]{article}

\usepackage[title]{appendix}

\usepackage{caption}
\usepackage{subcaption}
\usepackage{fullpage}
\usepackage{times}
\usepackage{array}
\usepackage{ragged2e}
\usepackage{multirow}
\usepackage{fancyhdr, amssymb, mathtools}
\usepackage[ruled,vlined]{algorithm2e}
\usepackage[dvipsnames,table]{xcolor}
\usepackage[margin = 1 in]{geometry}
\usepackage{graphicx}
\usepackage{outlines}
\usepackage{amsmath}

\usepackage{graphicx}
\usepackage{dirtytalk}

\usepackage{multirow}
\usepackage{arydshln}

\usepackage[english]{babel} 
\usepackage{blindtext}
\usepackage{amsfonts}
\usepackage{amssymb}
\usepackage{bbm}
\usepackage{mathtools}

\usepackage{hhline}
\usepackage{makecell}

\usepackage{xltabular}
\usepackage{arydshln} 
\usepackage{wrapfig}
\usepackage{enumitem}
\usepackage{floatrow}
\usepackage{xfrac}
\usepackage[font={footnotesize}, labelfont={bf}, labelsep=space, justification=justified, skip=0pt]{caption}
\usepackage[hang,flushmargin]{footmisc} 
\usepackage[colorlinks=true, allcolors=blue]{hyperref}
\usepackage{booktabs}
\usepackage{csquotes}

\definecolor{headcolor}{RGB}{255, 255, 255}
\definecolor{columncolor}{RGB}{255, 255, 255}
\definecolor{modelcolor}{RGB}{255, 255, 255}
\definecolor{optioncolor}{RGB}{255, 255, 255}

\definecolor{codegreen}{rgb}{0,0.6,0}
\definecolor{codegray}{rgb}{0.5,0.5,0.5}
\definecolor{codepurple}{rgb}{0.58,0,0.82}
\definecolor{backcolour}{rgb}{0.95,0.95,0.92}
\usepackage{siunitx}
\usepackage{listings}
\lstdefinestyle{mystyle}{
    backgroundcolor=\color{backcolour},   
    commentstyle=\color{codegreen},
    keywordstyle=\color{magenta},
    numberstyle=\tiny\color{codegray},
    stringstyle=\color{codepurple},
    basicstyle=\ttfamily\footnotesize,
    breakatwhitespace=false,         
    breaklines=true,                 
    captionpos=b,                    
    keepspaces=true,                 
    numbers=left,                    
    numbersep=5pt,                  
    showspaces=false,                
    showstringspaces=false,
    showtabs=false,                  
    tabsize=2
}
\usepackage{setspace}
\usepackage[numbers,sort&compress]{natbib}
\usepackage{cleveref} 
\AtBeginEnvironment{appendices}{\crefalias{section}{appendix}}
\newcommand{\cmt}[1]{} 
\newcommand{\ie}{i.e., }
\newcommand{\eg}{e.g., }

\newcommand{\xb}{\boldsymbol{x}}

\newcommand{\zb}{\boldsymbol{z}}

\newcommand{\tb}{\boldsymbol{t}}

\newcommand{\thetab}{\boldsymbol{\theta}}

\newcommand{\loss}{\mathcal{L}}

\newcommand{\yhb}{\boldsymbol{y}^{s_0}}
\newcommand{\yhbhat}{\hat{\boldsymbol{y}}^{s_0}}

\newcommand\ysbindout[2]{{y}^{s_{#1}}_{#2}}
\newcommand\ysbindoutk[2]{{y}^{s_{#1}(k)}_{#2}}

\newcommand\brackets[1]{\left[#1\right]}
\newcommand\parens[1]{\left(#1\right)}
\newcommand\braces[1]{\left\{#1\right\}}

\newcommand{\yhout}[1]{y^{s_0}_{#1} \big( \boldsymbol{x} \big)}

\newcommand{\ysout}[2]{y^{s_{#1}}_{#2}}

\newcommand{\thetabshat}[1]{\hat{\boldsymbol{\theta}}^{s_{#1}}}
\newcommand{\thetashatind}[2]{\hat{\boldsymbol{\theta}}^{s_{#1}}_{#2}}
\newcommand{\yhhatthetagenshat}[1]{%
    \hat{y}^{s_0}_{#1} \big( \boldsymbol{x}, \hat{\boldsymbol{\theta}}^s ; \hat{\boldsymbol{\phi}} \big)%
}

\newcommand{\yshatthetashat}[2]{%
    \hat{y}^{s_{#2}}_{#1} \big( \boldsymbol{x}, \hat{\boldsymbol{\theta}}^{s_{#2}}; \hat{\boldsymbol{\phi}} \big)%
}

\newcommand{\ybshat}[1]{%
    \hat{\boldsymbol{y}}^{s_{#1}}
}

\newcommand{\muhatnos}[1]{\hat{\mu}_{\theta_{#1}}}
\newcommand{\sigmahatnos}[1]{\hat{\sigma}_{\theta_{#1}}}
\newcommand{\thetabhatmse}{\hat{\boldsymbol{\theta}}^{MSE}}

\newcommand{\fullname}{Interpretable Probabilistic Neural Calibration}
\newcommand{\name}{iPro-NC}
\newcommand{\sourceind}[1]{s_{#1}}
\newcommand{\sourceinp}[1]{t_{\sourceind{#1}}}

\newcommand{\catind}{c}
\newcommand{\catinp}{\boldsymbol{t}_{\catind}}
\newcommand{\catinps}[1]{\boldsymbol{t}_{\catind}^{s_{#1}}}
\newcommand{\numinp}{\boldsymbol{x}}
\newcommand{\calind}{\theta}
\newcommand{\calinp}{\boldsymbol{\theta}}
\newcommand{\calest}[1]{\hat{\boldsymbol{\theta}}^{s_{#1}}}

\newcommand{\zblock}[1]{\boldsymbol{z}^{#1}}

\newcommand{\zcat}{\zblock{\catind}}
\newcommand{\zcal}[1]{\zblock{\calind_{#1}}}
\newcommand{\OHenc}[1]{\zeta\parens{#1}}
\newcommand{\fullout}[0]{\boldsymbol{\varUpsilon}}
\newcommand{\indout}[1]{\boldsymbol{y}_{#1}}
\newcommand{\fulloutmean}[0]{\hat{\boldsymbol{\mu}}_{\fullout{}}}
\newcommand{\fulloutstd}[0]{\hat{\boldsymbol{\sigma}}_{\fullout{}}}
\newcommand{\indoutmean}[1]{\hat{\boldsymbol{\mu}}_{\indout{#1}}}
\newcommand{\indoutstd}[1]{\hat{\boldsymbol{\sigma}}_{\indout{#1}}}
\newcommand{\calmean}[2]{\hat{\boldsymbol{\mu}}_{\calind{#1}}^{#2}}
\newcommand{\calstd}[2]{\hat{\boldsymbol{\sigma}}_{\calind{#1}}^{#2}}
\newcommand{\priorstd}{{\sigma}_{p}}

\newcommand{\lbnd}[1]{\hat{\boldsymbol{l}}^{#1}}
\newcommand{\ubnd}[1]{\hat{\boldsymbol{u}}^{#1}}
\newcommand{\bigone}[1]{\mathbbm{1} \braces{#1}}

\newcommand{\ExIIname}[0]{\text{Analytic Example}}

\newcommand{\jointsym}{\protect\footnotemark[2]}
\newcommand{\corrsym}{\protect\footnotemark[1]}
    \makeatletter
\def\@fnsymbol#1{\ensuremath{\ifcase#1\or \dagger\or *\or \ddagger\or
   \mathsection\or \mathparagraph\or \|\or **\or \dagger\dagger
   \or \ddagger\ddagger \else\@ctrerr\fi}}
    \makeatother

\usepackage{titling}
\thanksheadextra{}{}
\thanksheadextra{}{} 
\usepackage{lipsum} 

\title{Should We Simultaneously Calibrate Multiple Computer Models?}
\date{\vspace{-5ex}}
\usepackage{authblk}
\author[1]{Jonathan Tammer Eweis-Labolle\jointsym}
\author[1]{Tyler Johnson\jointsym}
\author[1]{Xiangyu Sun}
\author[1,2]{Ramin Bostanabad\corrsym}  
\affil[1]{Department of Mechanical and Aerospace Engineering, University of California, Irvine}
\affil[2]{Department of Civil and Environmental Engineering, University of California, Irvine}

\begin{document}

    \pagenumbering{arabic}
    \sloppy
    \maketitle
    
    \begingroup
    \renewcommand{\thefootnote}{}
    \footnotetext{\textsuperscript{*}Joint First Authors}
    \footnotetext{\textsuperscript{†}Corresponding Author: Raminb@uci.edu}
    \endgroup
    \section*{Abstract}
In an increasing number of applications designers have access to multiple computer models which typically have different levels of fidelity and cost. Traditionally, designers calibrate these models one at a time against some high-fidelity data (e.g., experiments). In this paper, we question this tradition and assess the potential of calibrating multiple computer models at the same time. To this end, we develop a probabilistic framework that is founded on customized neural networks (NNs) that are designed to calibrate an arbitrary number of computer models. In our approach, we (1) consider the fact that most computer models are multi-response and that the number and nature of calibration parameters may change across the models, and (2) learn a unique probability distribution for each calibration parameter of each computer model, (3) develop a loss function that enables our NN to emulate all data sources while calibrating the computer models, and (4) aim to learn a visualizable latent space where model-form errors can be identified. 
We test the performance of our approach on analytic and engineering problems to understand the potential advantages and pitfalls in simultaneous calibration of multiple computer models. Our method can improve predictive accuracy, however, it is prone to non-identifiability issues in higher-dimensional input spaces that are normally constrained by underlying physics.

\noindent \textbf{Keywords:} Model Calibration; Multi-fidelity Modeling; Uncertainty Quantification; Probabilistic Neural Networks; Inverse Problems; Manifold Learning; Data Fusion. 

    \section{Introduction} \label{sec introduction}

Computer models are increasingly employed in the design of complex systems for which direct observations may be scarce \cite{sacks_design_1989, forest_inferring_2008, salter_uncertainty_2019, larssen_forecasting_2006, arhonditsis_eutrophication_2007, henderson_bayesian_2009, gattiker_combining_2006}. Such models are typically built to simulate a range of expected behaviors \textemdash{} for instance, a given finite element (FE) model can simulate a wide variety of materials, loading conditions, failure modes, and so on \cite{baltic_2021_MLFractureLocusCalibration}. This flexibility typically relies on having \textit{calibration parameters} which must be tuned (i.e., calibrated) against some experimental or observational data such that the model's output matches the behavior of the true system as closely as possible before the model is used in design. 

To contextualize the calibration problem and some of its challenges, consider simulating the tension test (with fracture) of a metallic alloy via the FE method. There are a number of constitutive laws available such as Gurson-Tvergaard-Needleman (GTN) and J2 plasticity which can be used to model the material behavior. These laws not only have varying degrees of accuracy and cost, but also have shared and disparate tuning parameters \cite{abaqus_documentation}. While parameters such as Young's modulus, Poisson's ratio, and yield stress all have physical significance and exist in both laws, others account for issues such as mesh-dependency or missing physics and hence are specific to each constitutive law. Hence, in this problem, our goal is to match limited experimental data by \textit{calibrating} the parameters of each constitutive law, i.e., we want the resulting FE models to perform well and avoid overfitting. 

As reviewed in Section~\ref{sec related work}, existing calibration approaches have a few common characteristics. For instance, they all rely on surrogate models such as Gaussian processes (GPs) or neural networks (NNs) to reduce the data collection costs while increasing the efficiency of exploring (or sampling from) the parameter search space. In the relevant literature, the target/true system is typically referred to as the high-fidelity (HF) source while computer models (which require calibration) are generally considered as low-fidelity (LF) sources. With this terminology in mind, we note that existing methods almost always work in a bi-fidelity setting where a single LF source is calibrated using limited samples from an HF source. Given the availability of multiple LF sources in most engineering applications, a natural question arises: Can we calibrate these LF sources simultaneously?

Answering this question is challenged by the fact that LF sources typically have $(1)$ a different number of calibration parameters which may or may not have physical meanings, $(2)$ varying degrees of fidelity where it is generally more costly to sample from the more accurate LF sources, and $(3)$ multiple responses. 
Our goal in this paper is to design a framework that simultaneously calibrates multiple LF sources regardless of their accuracy/fidelity levels or the number/nature of their calibration parameters. We build our framework via NNs and use it to study the potential benefits and drawbacks of simultaneous calibration of multiple LF sources.

The rest of the paper is organized as follows:
We provide some literature review in Section \ref{sec related work} and introduce our approach in Section \ref{sec method}. We assess the performance of our approach on two examples in Section \ref{sec results} and then provide further discussions and concluding remarks in Sections \ref{sec discussion} and \ref{sec conclusion}, respectively.
    \section{Related Works} \label{sec related work}

\noindent \textbf{Gaussian Processes:} Many calibration frameworks leverage GPs because they are probabilistic surrogates (i.e., emulators) that are easy to fit, interpretable, and robust to overfitting \cite{Rasmussen2006Gaussian, eweis2022data, deng2023data, yousefpour_gp_2023}. GPs provide priors over function spaces by assuming that the observations follow a multivariate normal distribution (MVN) whose mean vector and covariance matrix are constructed via the GP's parametric mean function and kernel (aka covariance function). Given some input-output pairs, a GP can be easily trained via maximum likelihood estimation (MLE) \cite{planas_extrapolation_2020}.

\noindent \textbf{GP-based Calibration:} 
GPs can be used for calibration either directly or as an integral part of structured frameworks. A prominent example of the latter is the work of Kennedy and O'Hagan (KOH) \cite{kennedy2001bayesian} who leverage GPs to additively relate a computer model to the corresponding physical system:
\begin{equation}
    \eta^{h}{\parens{\numinp{}}} = \eta^{l}{\parens{\numinp{}, \thetab{}^*}} + \delta{\parens{\numinp{}}}
    \label{eqn: ProNDFCal_KOH}
\end{equation}
where $\numinp{}$ are the system inputs, $\thetab{}^*$ are the true calibration parameters, and $\eta^{l}(\cdot)$, $\eta^{h}{}(\cdot)$, and $\delta(\cdot)$ are GP emulators of the LF source, HF source, and discrepancy/bias function, respectively. 
The parameters of these GPs as well as $\thetab$ can be estimated via fully \cite{higdon2004combining, plumlee2017bayesian} or modular \cite{zhang2019numerical, apley2006understanding, bayarri2007framework, arendt2012quantification, arendt2012improving} Bayesian inference. 
Although this method has been successfully applied to many applications \cite{stainforth2005uncertainty, gramacy2015calibrating,zhang2019numerical}, it only accommodates bi-fidelity problems, suffers from identifiability issues, scales poorly to high-dimensional problems, and imposes an additive nature on the bias \cite{kennedy2001bayesian, higdon2004combining, RN648, RN1071}. 
KOH's method has been extended in a number of important directions that address some of these issues by, e.g., using multi-response data to reduce non-identifiability issues \cite{RN271, RN664, arendt2012improving} or by using principal component analysis to enable scalability to high-dimensional outputs \cite{higdon_HDO_2008}. 

Recent works have used GPs directly for calibration by reformulating their kernel \cite{eweis2022data, yousefpour_gp_2023}. The main idea of these works is to augment the input space via a categorical source-indicator variable that is internally converted to some quantitative latent variables which are then passed to the kernel. The addition of this categorical variable also allows for the direct inclusion of calibration parameters in the kernel as unknown variables which are estimated either deterministically or probabilistically during training. This approach does not place any assumption on the form of the bias (e.g., additive or multiplicative) and works with an arbitrary number of LF sources as long as they share the same calibration parameters. Being a GP-based approach, however, does impose some limitations on this method too.

\noindent \textbf{NN-based Calibration:} 
NNs offer a number of advantages over GPs such as scalability to large data and high dimensions. NNs can be trained in a variety of ways with a wide range of available architectures, activation functions, and loss functions. They can also incorporate probabilistic elements through, \eg dropout layers \cite{10.5555/3045390.3045502,RN1074}, variational formulations \cite{RN1191}, or Bayesian treatments \cite{blundell2015weight}. However, in the absence of abundant data (as is often the case in calibration problems), NNs' performance are highly sensitive to these modeling choices \cite{lecun2015deep} which necessitates extensive and costly tuning of their architecture and parameters \cite{eweis2022data}. As such, NN-based calibration approaches primarily aim to alleviate these issues by, e.g., reducing the scope of the problem or incorporating domain knowledge.
For example, \cite{baltic_2021_MLFractureLocusCalibration} casts calibration as a forward problem where a convolutional neural network (CNN) maps the outputs to the parameter space. Upon training on LF data, the NN is fed HF data to estimate the calibration parameters. Forward techniques are straightforward and can incorporate a variety of data types but they struggle with highly nonlinear systems and fail to capture the bias between LF and HF sources. Other recent works have explored problems governed by partial differential equations (PDEs) whose parameters are estimated such that the PDE solution matches observational data \cite{hamel_calibrating_2023, yousefpour_GPPINNs_2024}.
    \section{Proposed Approach} \label{sec method}

Our goal is to design a calibration framework that addresses the following five major challenges.\\
\textbf{(1) Number and dimensionality of LF sources:} Multiple ($\geq2$) LF sources may be available which can have distinct and/or shared calibration parameters.\\
\textbf{(2) Bias and noise:} The relationships between the LF and HF sources may be unknown and each data source may be corrupted by noise of different variance.\\
\textbf{(3) Data imbalances:} LF samples dominate the size of the dataset since LF sources are typically much cheaper to query.\\
\textbf{(4) Output dimensionality:} The number of outputs or their dimensionality can be high (e.g., if a response is an image or curve) and the dependence on the calibration parameters can dramatically vary across different outputs.\\
\textbf{(5) Uncertainty sources:} There are multiple coexisting uncertainty sources which render parameter estimation sensitive to many factors such as dataset size or dimensionality. This feature makes it necessary to estimate the calibration parameters probabilistically.

To collectively address the above challenges, we propose \fullname{} (\name{}) which is a customized NN with a multi-block architecture 
that converts MF modeling and calibration to a latent variable modeling problem. 
To explain our rationale for taking this approach we note that the relation between data sources is unknown and can be very complex. This relation may change across different responses in multi-output applications and strongly depends on the calibration parameters whose dimensionality may depend on the data source. We argue that learning such complex relations is possible only via latent variables and \name{} is indeed designed to learn such variables. 
In addition to a custom architecture, \name{} has a novel loss function that is particularly designed for calibration problems, and it also provides visualizable latent variables that reveal the learned relationship between the data sources. 

Below, the matrix $\fullout{}$ contains all the outputs $\indout{i}$, \ie $\fullout{} = \brackets{\indout{1}, \indout{2}, \cdots, \indout{n_{\indout{}}}}$ where $n_{\indout{}}$ is the number of outputs and $d \fullout{} = \sum_{i}^{n_{\indout{}}}{d \indout{i}}$ where $d \indout{i}$ is the dimensionality of the $i^{th}$ output. 
Additionally, we denote numerical inputs via $\xb$, categorical variables (if any) via $\tb_c$, and the calibration parameters via $\thetab$. 


\subsection{Architecture: Information Flow} \label{subsec method_archictecture}

\begin{figure*}[!ht] 
    \centering
    \includegraphics[width = 1.15\textwidth, trim=70 0 0 10, clip]{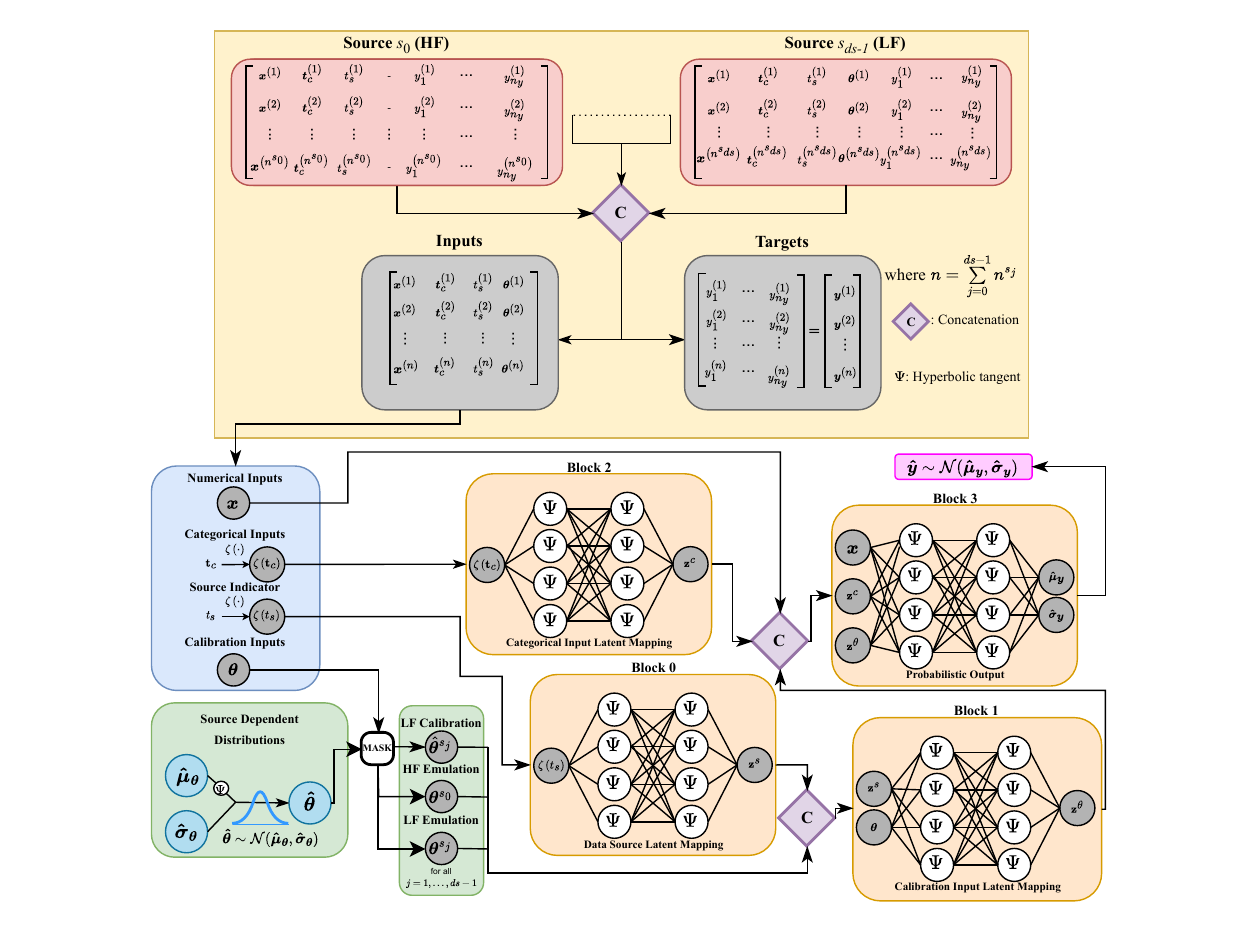}
    \vspace{-0.3cm}
    \caption{\textbf{\fullname{} (\name{}):} The proposed multi-block architecture allows us to fuse all sources of data and simultaneously calibrate an arbitrary number of LF models by estimating a unique set of distribution parameters for each LF source.}
    \label{fig: ProNDFCal_architecture}
\end{figure*}

Calibration requires MF modeling and so we design our network based on this dependence. 
As shown in Figure~\ref{fig: ProNDFCal_architecture} and motivated by \cite{mora_eweis2023prondf}, we convert MF modeling to a latent variable learning problem (we adopt this approach because it allows us to handle all five challenges associated with calibration).
This conversion is achieved by augmenting the input space with the categorical variable $t_s$ which simply denotes the source of a sample. $t_s$ is categorical and hence agnostic to the order and fidelity level of the data sources but for notational simplicity we consider the HF source as source zero, i.e., $\sourceind{0}$. 

Once $t_s$ is added, we concatenate all the datasets and pass the inputs to the network which processes $\xb$, $\tb_c$, $\thetab$, and $t_s$ differently. 
Specifically, \name{} learns quantitative embeddings for $\tb_c$ and $t_s$ by first one-hot encoding them via the deterministic functions $\zeta(\tb_c)$ and $\zeta(t_s)$, respectively, and then passing them through Blocks 2 and 0 which are NNs with low-dimensional outputs $\zb^c$ and $\zb^s$. The latent variable $\zb^s$ represents the relationships between the different sources of data, while $\zb^c$ represents the relationships between categorical combinations. 
The latent variables learned for $\tb_c$, i.e., $\zb^c$, are now ready to be combined with numerical inputs $\xb$ but $\zb^s$ need some additional work before they can be combined with $\zb^c$ and $\xb$: 
to capture the effect of the data source on the calibration parameters, we first concatenate $\zb^s$ with \textit{masked} calibration inputs and then map the combined vector via Block 1 to latent variable $\zcal{}$ which can now represent the effects of the tuning parameters on the LF sources.

The reason we use masking is that the calibration parameters that are used for an LF source depend on whether \name{} is tasked to (1) emulate the HF source with that LF source, or (2) emulate the LF source itself. For the former case, each forward pass in the model leverages samples from a multivariate normal distribution that is unique to an LF source. However, for the latter case, the calibration parameters in the LF data are used. Masking also allows us to consider the fact that the HF source does not have any calibration parameters. In this case we use some dummy values for $\thetab$ and note that the output of \name{} should be insensitive to these dummy values (we achieve this insensitivity by adding a term to the loss function).

Once all latent variables are obtained and concatenate with $\xb$, they are fed into Block 3 which outputs a normal distribution for each output. With this setup, the network can be trained via \textit{all} available data at once.

\subsubsection{Interpretability for Decision Making} \label{subsec method_interpret}
By design, our multi-block architecture distributes the various tasks to different blocks of the network to not only aid in learning, but also provide interpretable metrics that facilitate decision making. 
For instance, Block 0 takes as input $\zeta(t_s)$ which is a deterministic encoding of $t_s$ and outputs a learned embedding that visualizes the relationships between data sources. Data sources that are recognized to have similar input-output patterns are encoded close-by, while those with less correlation are more distant. Unlike most MF modeling and calibration works such as that of KOH, this approach does not impose any \textit{a priori} relation between any of the sources. 


The relations between the data sources depends on the estimated calibration parameters for each source, i.e., $\calest{i}$ for $i=1, \cdots, ds$. \name{} uses Block 1 to capture this dependence by combining $\zb^s$ with the masked calibration parameters and then passing the combined vector through a few hidden layers to obtain $\zcal{}$. 
Based on our network architecture, we can expect specific trends in the $\zcal{}$ space. For example, when \name{} is used to emulate the HF source by setting $t_s=\sourceind{0}$, the dummy values used as the calibration parameters of the HF source should not affect $\zcal{}$ since the HF source does not have any calibration parameters. That is, once \name{} is trained, if we set $t_s=\sourceind{0}$ during inference, we should see a compact distribution in $\zcal{}$.

Unlike the $t_s=\sourceind{0}$ case, when $t_s\neq\sourceind{0}$ we expect to see a distribution of points in the $\zcal{}$ space for each LF source. For the $i^{th}$ LF source, this distribution depends on $\calest{i}$, which is modeled via a multivariate normal distribution whose mean vector and covariance matrix are learned during training. 
Our estimated calibration parameters should be found within the range of our sampled $\calinp{}$. These calibration parameters should minimize the distance between the points in $\zcal{}$. This distance $\left\| \zcal{\sourceind{0}} - \zcal{\sourceind{i}}(\calest{i}) \right\|$ provides a direct measure of how accurate the calibrated model is relative to the HF source.

Block 2 serves a similar purpose as Block 0, except that it reveals the relationships between the combinations of the levels of the categorical variables $\catinp$ present in the data (if any) by mapping them to $\zcat{}$.

Finally, Block 3 learns the effects of $\xb$ and the learned latent variables on the outputs where each output is modeled as a normal distribution, i.e., the network outputs the two vectors $\fulloutmean{}$ and $\fulloutstd{}$ which are the parameters of independent normal distributions \cite{10.5555/3495724.3496975}.
The reason for modeling the outputs as distributions is to quantify the aleatoric uncertainties especially in cases where the noise variance changes across different sources. 
Another benefit of using distributions is the ability to incorporate a proper scoring rule \cite{mora_eweis2023prondf} into our loss function, which provides more accurate prediction intervals and helps prevent overfitting (see \ref{subsec method_loss}). 

\subsection{Loss Function: Emulation and Calibration} \label{subsec method_loss}

Our loss has multiple terms since \name{} aims to emulate the input-output relationship of each data source, estimate the \textit{optimal} calibration parameters for each LF source, \ie those which minimize the error with respect to the HF source, learn from unbalanced and scarce data without over-fitting, and ensure that the tasks for each system output are learned at roughly the same rates.
Our loss is defined as:
\begin{equation}
\begin{split}
    \loss = 
    \sum_{i=1}^{n_{\indout{}}} &\Bigg(
     \loss_{NLL_i}^{em}
    + \loss_{NLL_{i}}^{cal} \Bigg)
    + \beta_{IS} \parens{\loss_{IS}^{em}
    + \loss_{IS}^{cal}}
    + \beta_{KL} \loss_{KL}
    \label{eqn: ProNDFCal_loss}
\end{split}
\end{equation}
\noindent where $\loss_{NLL}$ refers to the negative log likelihood, $\loss_{IS}$ refers to the interval score (see \cite{mora_eweis2023prondf}), $\loss_{KL}$ is regularization term based on the KL-divergence, $\beta_{IS}$ and $\beta_{KL}$ are tunable hyperparameters which weight their respective loss components, $em$ denotes an emulation task, $cal$ denotes a calibration task, and $n_{\indout{}}$ is the number of system outputs. 
Note that the loss does not include the typical $L2$ regularization as we employ weight decay on the network parameters (excluding the parameters representing the calibration estimates) via the Adam optimizer \cite{loshchilov_decoupled_2019} (see \ref{sec discussion} for further discussion on regularization).
We consider the divergence only between the standard deviations since the mean should be found based on the true value of the calibration parameters (which may be unknown).

The individual loss terms in \ref{eqn: ProNDFCal_loss} are calculated as follows:
{\footnotesize
\begin{align}
\loss_{NLL_i}^{em} &= \sum_{j=0}^{d \sourceind{} - 1} \loss_{NLL} \Bigg\{
    \indout{i}^{\sourceind{j}},
    \indoutmean{i} \parens{
    \OHenc{\sourceinp{} = \sourceind{j}},
    \calinp^{\sourceind{j}},
    \OHenc{\catinps{j}},
    \numinp^{\sourceind{j}}}, 
    \indoutstd{i} \parens{
    \OHenc{\sourceinp{} = \sourceind{j}},
    \calinp^{\sourceind{j}},
    \OHenc{\catinps{j}},
    \numinp^{\sourceind{j}}}\Bigg\}
    \label{eqn: ProNDFCal_lossNLLem}
    \\[2.5ex]
\loss_{NLL_{i}}^{cal} &= \sum_{j=0}^{d \sourceind{} - 1} \loss_{NLL} \Bigg\{
    \indout{i}^{\sourceind{0}},
    \indoutmean{i} \parens{
    \OHenc{\sourceinp{} = \sourceind{j}},
    \calest{j},
    \OHenc{\catinps{j}},
    \numinp^{\sourceind{j}}}, 
    \indoutstd{i} \parens{
    \OHenc{\sourceinp{} = \sourceind{j}},
    \calest{j},
    \OHenc{\catinps{j}},
    \numinp^{\sourceind{j}}}\Bigg\}
    \label{eqn: ProNDFCal_lossNLLcal}
    \\[2.5ex]
    \loss_{IS}^{em} &= \loss_{IS} \Bigg\{
    \indout{},
    \indoutmean{} \parens{
    \OHenc{\sourceinp{} = \sourceind{j}},
    \calinp,
    \OHenc{\catinp},
    \numinp}, 
    \indoutstd{} \parens{
    \OHenc{\sourceinp{} = \sourceind{j}},
    \calinp,
    \OHenc{\catinp},
    \numinp}\Bigg\}
    \label{eqn: ProNDFCal_lossISem}
    \\[2.5ex]
    \loss_{IS}^{cal} &= \sum_{j=1}^{d \sourceind{} - 1} \loss_{IS} \Bigg\{
    \indout{}^{\sourceind{0}},
    \indoutmean{} \parens{
    \OHenc{\sourceinp{} = \sourceind{j}},
    \calest{j},
    \OHenc{\catinps{j}},
    \numinp^{\sourceind{j}}},
    \indoutstd{} \parens{
    \OHenc{\sourceinp{} = \sourceind{j}},
    \calest{j},
    \OHenc{\catinps{j}},
    \numinp^{\sourceind{j}}}\Bigg\}
    \label{eqn: ProNDFCal_lossIScal}
    \\[2.5ex]
    \loss_{NLL} \big\{\indout{}, \indoutmean{}, \indoutstd{}\big\} &= 
    - \frac{1}{N} \sum_{k=1}^{N} \log \mathcal{N} \parens{
    \indout{}^{(k)}; \indoutmean{}^{(k)}, \parens{\indoutstd{}^{(k)}}^2}\!\!
    \label{eqn: ProNDFCal_lossNLL} 
    \\[2.5ex]
    \loss_{IS} \big\{\indout{}, \indoutmean{}, \indoutstd{}\big\} &= 
    \frac{1}{N} \sum_{k=1}^{N} \Bigg[
    \parens{\ubnd{(k)} - \lbnd{(k)}}
    +
    \frac{2}{\varphi}
    \parens{\lbnd{(k)} - \indout{}^{(k)}}
    \bigone{\indout{}^{(k)} < \lbnd{(k)}}
    + \frac{2}{\varphi}
    \parens{\indout{}^{(k)} - \ubnd{(k)}}
    \bigone{\indout{}^{(k)} > \ubnd{(k)}}
    \Bigg]
    \label{eqn: ProNDFCal_lossIS} 
    \\[2.5ex]
    \loss_{KL} \big\{\calstd{}{} \big\} &= 
    \sum_{i=0}^{d\calind{} - 1} \sum_{j=0}^{d \sourceind{} - 1} \Bigg(
     \log \parens{\frac{\calstd{i}{\sourceind{j}}}{\priorstd}}
      + \frac{\priorstd}{2\calstd{i}{\sourceind{j}}} - 0.5 \Bigg)
    \label{eqn: ProNDFCal_lossKL}
\end{align}
}
\noindent where $\indout{i}^{\sourceind{j}}$ is the $i^{th}$ output of source $j$, $\indoutmean{i}$ and $\indoutstd{i}$ are, respectively, the network predictions for the means and standard deviations for the $i^{th}$ output.
$\sourceinp{} = \sourceind{j}$, $\calinp^{\sourceind{j}}$, $\catinps{j}$, and $\numinp^{\sourceind{j}}$ are, respectively, the source inputs, calibration inputs, categorical inputs, and numeric inputs for the $j^{th}$ data source, $\calest{j}$ are the estimated calibration parameters for the $j^{th}$ source (sampled via the reparameterization trick, and where $j \neq 0$, \ie an LF source), $\OHenc{\cdot}$ represents a one-hot encoding, $ds$ is the number of data sources (note that the data sources are indexed from $0$, \eg for three sources we have $[\sourceind{0}, \sourceind{1}, \sourceind{2}]$ with $ds=3$), $N$ is the number of samples (\eg in a training batch), $(k)$ denotes an individual sample, $\bigone{\cdot}$ is an indicator function that returns $1$ if the event in brackets is true and $0$ otherwise, $\calstd{i}{\sourceind{j}}$ is the parameter representing the standard deviation for the $i^{th}$ calibration parameter for the $j^{th}$ LF source, and $\priorstd$ is the prior for the standard deviation (which should either be set based on domain knowledge or tuned). 

We highlight that to ensure that the network learns equally from each source in spite of a possible large data imbalance, $\loss_{NLL_i}^{em}$ is calculated by separately calculating $\loss_{NLL_i}$ for each source, normalizing by the number of samples for that source in the batch, and then summing the results, promoting each data source to equally contribute to the loss. 

The loss terms in Equation \ref{eqn: ProNDFCal_loss} are of four types, i.e.,  $\loss_{NLL_i}^{em}$, $\loss_{NLL_i}^{cal}$, $\loss_{IS_i}^{em}$, and $\loss_{IS_i}^{cal}$.
The likelihood terms $\loss_{NLL_i}^{em}$ and $\loss_{NLL_i}^{cal}$ penalize the model if the training data is unlikely to have been generated by the predicted distributions.
The interval score terms $\loss_{IS_i}^{em}$ and $\loss_{IS_i}^{cal}$ reward narrow prediction intervals (PIs) but penalize the model for each data point outside the $(1-\varphi)\times\text{100}\%$ PI spanning $\brackets{\lbnd{(k)}, \ubnd{(k)}}$ where $\lbnd{} = \indoutmean{} - 1.96 \indoutstd{}$ and $\ubnd{} = \indoutmean{} + 1.96 \indoutstd{}$. We use $\varphi = 5\%$, meaning that $\loss_{IS}$ is minimized by a distribution whose $95\%$ PI is as tight as possible while still containing all training samples.

The emulation terms $\loss_{NLL_i}^{em}$ and $\loss_{IS}^{em}$ encourage the model's predictions for output $i$ on each source of data to match the training data $\indout{i}^{\sourceind{j}}$ for the corresponding source. The former is obtained via distributions, i.e., by $\indoutmean{i} \parens{\sourceinp{} = \sourceind{j}, \calinp^{\sourceind{j}}, \catinps{j}, \numinp^{\sourceind{j}}}$ and $\indoutstd{i} \parens{\sourceinp{} = \sourceind{j}, \calinp^{\sourceind{j}}, \catinps{j}, \numinp^{\sourceind{j}}}$ for each source $\sourceind{j}$ with $j=0,1,2,\cdots,ds-1$,
We highlight that when missing elements of $\calinp{}$ are substituted with portions of $\calest{}$, we do not allow gradients on the emulation portion of the loss to back propagate to the model's estimated calibration parameters. This ensures that the emulation task does not interfere with the calibration task.

The calibration terms $\loss_{NLL_i}^{cal}$ and $\loss_{IS_j}^{cal}$ encourage the model's predictions for output $i$ on each \textit{LF source} with the estimated calibration parameters to match the training data $\indout{i}^{\sourceind{0}}$ for the \textit{HF source}. That is, we want the model to find calibration estimates which make the LF outputs best match the HF source. The model's predictions in this case are given by normal distributions with $\indoutmean{i} \parens{\sourceinp{} = \sourceind{j}, \calest{j}, \catinps{j}, \numinp^{\sourceind{j}}}$ and $\indoutstd{i} \parens{\sourceinp{} = \sourceind{j}, \calest{j}, \catinps{j}, \numinp^{\sourceind{j}}}$ for each LF source $\sourceind{j}$ with $j=1,2,\cdots,ds-1$.


\subsection{Training and Prediction} \label{subsec method_training}

Our model is composed of connected feed-forward blocks so training and prediction are relatively straightforward. However, the calibration parameters require some special treatment. While $\loss_{NLL}$ and $\loss_{IS}$ terms may be written directly as functions of the parameters of the output distribution, the same is not true for $\calmean{}{}$ and $\calstd{}{}$ as loss gradients cannot be back-propagated directly through parameterized distributions of this sort \cite{RN941}. 
So, we sample $\calest{}$ from $\calmean{}{}$ and $\calstd{}{}$ via the ``reparameterization trick'' \cite{RN941} which enables us to train the network directly with typical back-propagation. To prevent the network from extrapolating when estimating the distribution parameters, \ie sampling $\calest{}$ which lie outside of the training domain, we clamp $\calest{}$ to this domain via a scaled hyperbolic tangent activation function before retrieving them or passing them to Block 1 (we select hyperbolic tangent rather than another clamping function, \eg sigmoid, because it has larger gradients which aid in learning). This is essential because NN predictions are not trustworthy except in the regions spanned by the training data.
    \section{Results} \label{sec results}
We test our approach on an analytic example and an engineering problem in Sections \ref{subsec analytic_ex} and \ref{subsec engineering_prob}. 
We implement our approach using PyTorch Lightning \cite{PyTorch_Lightning} and train for $4,000$ (analytic example) or $14,000$ (engineering problem) epochs using a learning rate of $1\mathrm{e}{-2}$. In the analytic example, we use the entire available data in each batch since our dataset is quite small. We fix the architectures for Blocks 0, 1, and 2 to one hidden layer with $5$ neurons and the dimension of all manifolds to $2$. We fix the architecture for Block 3 as four hidden layers with 16, 32, 16, and 8 neurons and 32, 62, 32, and 16 neurons for the analytic and engineering examples, respectively. 
The feed-forward blocks are initialized randomly, while the calibration parameters are initialized to the mean of their distribution in the data. We use hyperbolic tangent as the activation function for all blocks and as such pre-process all numeric data via linear scaling to mean $0$ and standard deviation $1$. Our code, along with further implementation details, will be published on \href{https://github.com/Bostanabad-Research-Group}{GitHub} upon publication.

\subsection{Analytic Example} \label{subsec analytic_ex}
The analytic example is designed to test every goal that \name{} aims to achieve: it has three sources which have three responses and one numeric input, source $\sourceind{1}$ has two calibration parameter while $\sourceind{2}$ only has one calibration parameter, and $\sourceind{1}$ has model-form error (while $\sourceind{2}$ does not), and the HF source is corrupted by a different amount of noise on each output while the LF data are noise-free. 
The functional forms of the data sources are:

\allowdisplaybreaks
\begin{align}
    &\ysbindout{0}{1} = -0.5x^{3} - 2.0x^{2} + x + 1, \tag{10.1} \label{eq 10.1}\\ 
    &\ysbindout{0}{2} = \log\parens{-0.5x^{3} + 2.0x^{2} + 2.0x + 11}, \tag{10.2} \label{eq 10.2}\\ 
    &\ysbindout{0}{3} = -0.5x^{3} + 2.0\parens{x - 0.5}^{2} - 2 \tag{10.3} \label{eq 10.3}\\[5pt]
    &\ysbindout{1}{1} = \calind{}^{s_1}_{1} x^{3} - \calind{}^{s_1}_{2} x^{2} + 2, \tag{11.1} \label{eq 11.1}\\
    &\ysbindout{1}{2} = \log\parens{\calind{}^{s_1}_{1} x^{3} + \calind{}^{s_1}_{2} x^{2} + 2.0x + 11}, \tag{11.2} \label{eq 11.2}\\ 
    &\ysbindout{1}{3} = \calind{}^{s_1}_{1} x^{3} + \calind{}^{s_1}_{2} \cosh\parens{x - 0.3} - 3.5 \tag{11.3} \label{eq 11.3}\\[5pt]
    &\ysbindout{2}{1} = \calind{}^{s_2}_{1} x^{3} - 2.0x^{2} + x + 1, \tag{12.1} \label{eq 12.1}\\
    &\ysbindout{2}{2} = \log\parens{\calind{}^{s_2}_{1} x^{3} + 2.0x^{2} + 2.0x + 11}, \tag{12.2} \label{eq 12.2}\\ 
    &\ysbindout{2}{3} = \calind{}^{s_2}_{1} x^{3} + 2.0\parens{x - 0.5}^{2} - 2 \tag{12.3} \label{eq 12.3}\\[5pt]
    &x \in [-1, 2.2], \quad \theta \in [-1, 2.2], \quad \sigma^{2,~s_0} = [0.025, 0.00005, 0.02] \nonumber
\end{align}

\setcounter{equation}{12}

We consider three scenarios in this example: training on $\sourceind{0}$ and $\sourceind{1}$, $\sourceind{0}$ and $\sourceind{2}$, and all sources. In all cases, we generate $n^{s_0}=40, \ n^{s_1}=200,$ and $n^{s_2}=100$ samples from each source for training, $n/4$ data points for validation (where $n$ is the training data for a given source in a given problem), and an additional $1,000$ test samples for each source.
\begin{figure*}[!h] 
    \centering
    \begin{minipage}[t]{0.667\textwidth}
        \centering
        \includegraphics[width=1.0\linewidth]{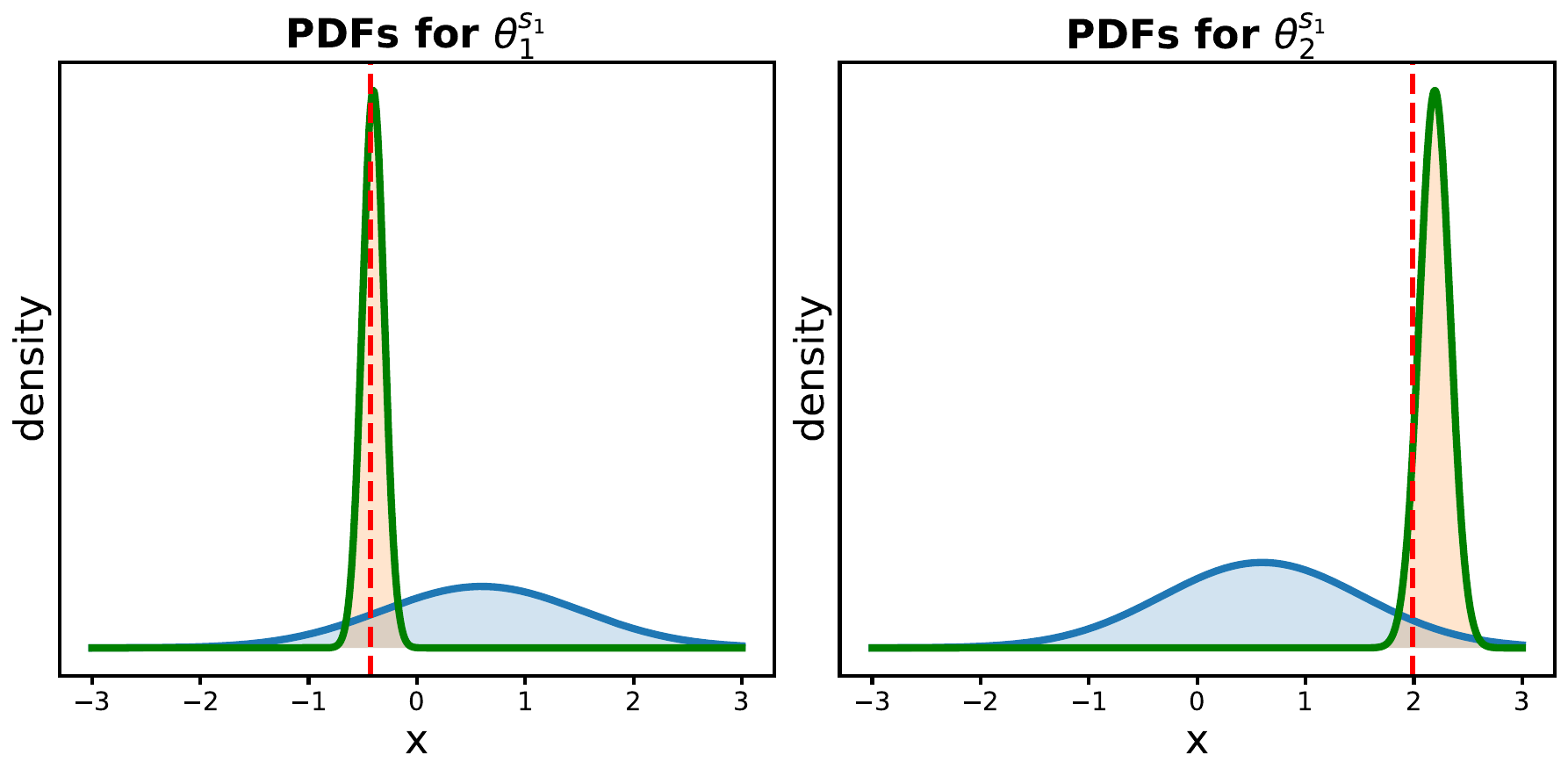}
        \vspace{-0.6cm}
        \subcaption{}
        \label{subfig iProNC_Analytic_CalDistLF1Only}
    \end{minipage}%
    \hfill 
    \begin{minipage}[t]{0.331\textwidth}
        \centering
        \includegraphics[width=1.0\linewidth]{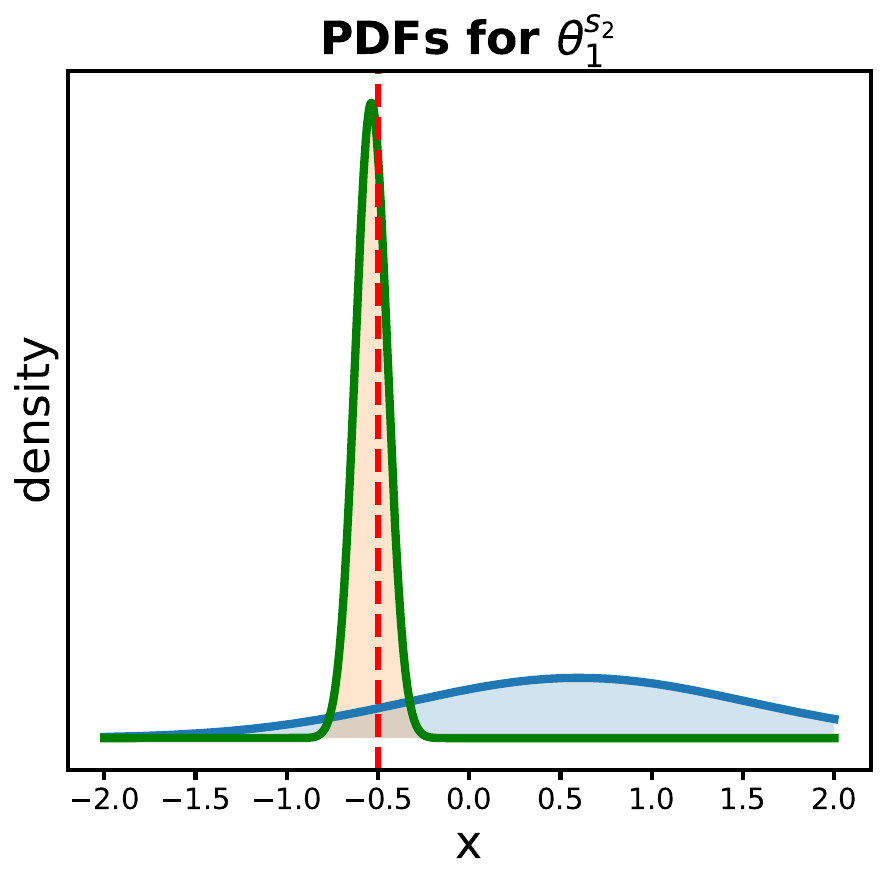}
        \vspace{-0.6cm}
        \subcaption{}
        \label{subfig iProNC_Analytic_CalDistLF2Only}
    \end{minipage}%
    \vfill 
    \begin{minipage}[t]{1.0\textwidth}
        \centering
        \includegraphics[width=1.0\linewidth]{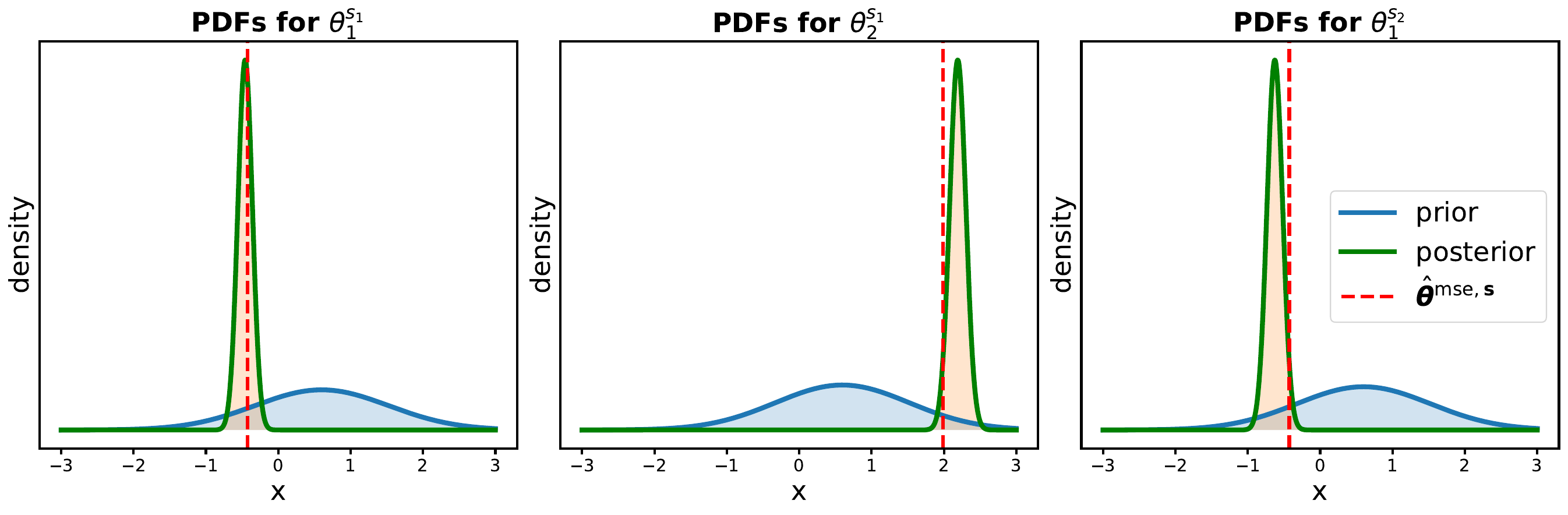}
        \vspace{-0.6cm}
        \subcaption{}
        \label{subfig iProNC_Analytic_CalDistAll}
    \end{minipage}
    \vspace{-0.3cm}
    \caption{\textbf{Calibration Inference for the \ExIIname:} (a) PDFs of $\thetashatind{1}{1}$ and $\thetashatind{1}{2}$ when trained on $\sourceind{0}$ and $\sourceind{1}$. (b) PDF of $\thetashatind{2}{1}$ when trained on $\sourceind{0}$ and $\sourceind{2}$.  (c) PDFs of $\thetashatind{j}{i}$ when trained on all sources.}
    \label{fig iProNC_Analytic_CalDist}
\end{figure*}
Table~\ref{tab iProNC_Analytic_HFTruePred} shows the accuracy in emulating each response of the HF source across the three scenarios when $t_s=s_0$ is used in \name{}. We observe that when all sources are included in the training, the network is never the worst performing on any output. This trend indicates that \name{} is effectively (1) leveraging data from all sources to more accurately emulate the HF source, and (2) removing the effect of dummy $\thetab$ on the predictions (recall that HF emulation with $t_s=s_0$ does not require calibration). 
\begin{table}[!h]
    \centering
    
    \begin{tabular}{|l|l|l|l|}
        \hline
        \textbf{Dataset} & \(\ysout{0}{1} \, \text{vs} \, \yhhatthetagenshat{1}\)
                & \(\ysout{0}{2} \, \text{vs} \, \yhhatthetagenshat{2}\)
                & \(\ysout{0}{3} \, \text{vs} \, \yhhatthetagenshat{3}\) \\
        \hline
        All Sources & \multicolumn{1}{c|}{0.0825}                      & \multicolumn{1}{c|}{0.0904}                      & \multicolumn{1}{c|}{0.1458}\\ \hline
        $\sourceind{0}$~and \ $\sourceind{1}$    & \multicolumn{1}{c|}{0.1043} & \multicolumn{1}{c|}{0.0877} & \multicolumn{1}{c|}{0.2045} \\ \hline
        $\sourceind{0}$~ and \ $\sourceind{2}$    & \multicolumn{1}{c|}{0.0677} & \multicolumn{1}{c|}{0.1144} & \multicolumn{1}{c|}{0.1785} \\ \hline
    \end{tabular}
    \caption{\textbf{Analytic Example:} RRMSE on emulation accuracy for \(\yhbhat\) (with estimated dummy calibration parameters) vs \(\yhb\).}
    \label{tab iProNC_Analytic_HFTruePred}
\end{table}
\begin{figure*}[!ht]
    \centering
    \begin{subfigure}{0.83\textwidth}
        \centering
        \includegraphics[width=1.0\linewidth, trim=6 10 7 5, clip]{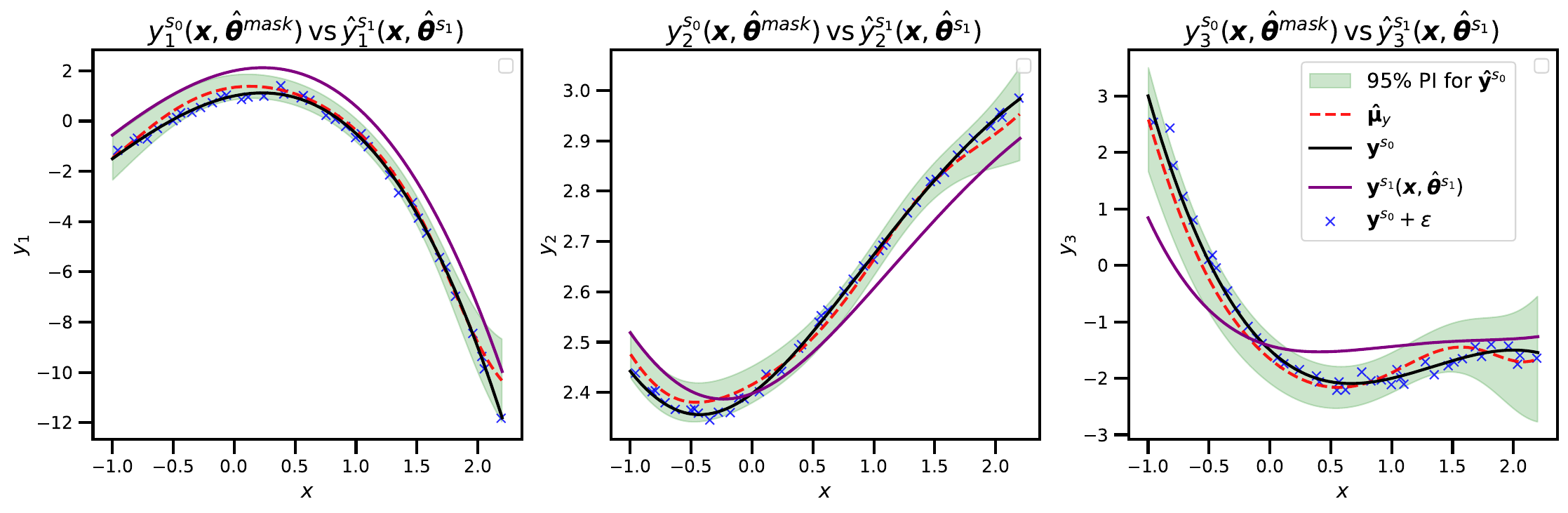}
        \caption{$\indout{}^{\sourceind{0}}$ true vs $\sourceind{1}$ via $\calest{1}$ using 2 Sources}
        \label{subfig iProNC_Analytic_HFPredEstCalsLF1Only}
    \end{subfigure}%
    \hfill
    \begin{subfigure}{0.83\textwidth}
        \centering
        \includegraphics[width=1.0\linewidth]{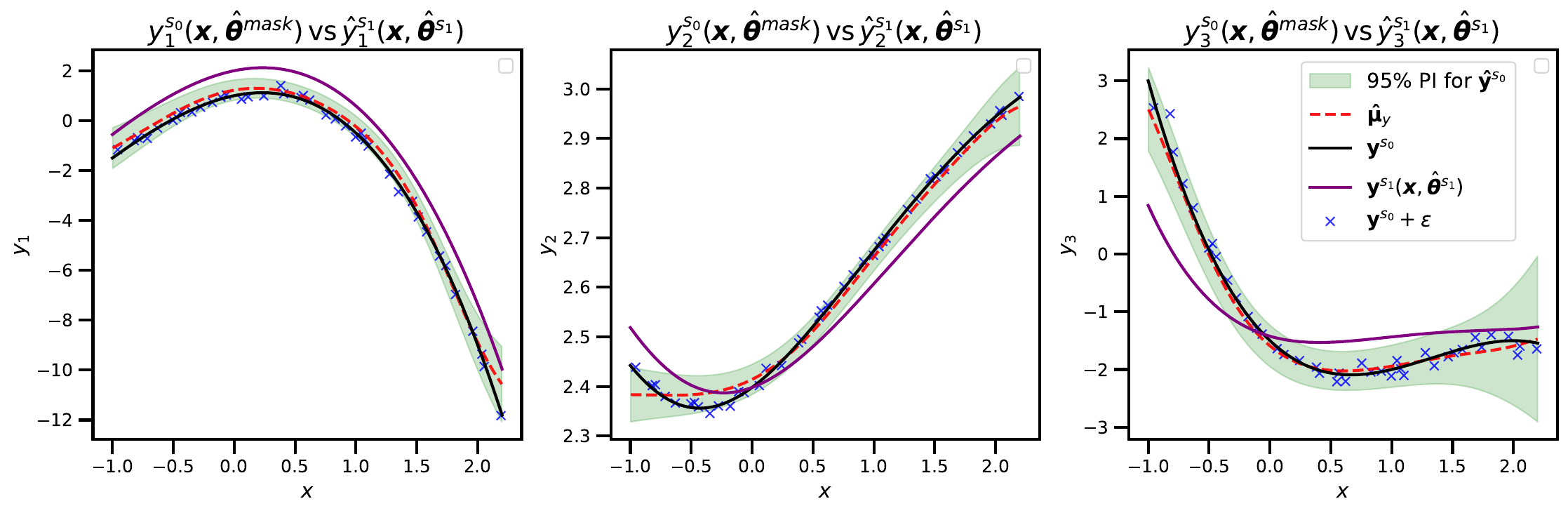}
        \caption{$\indout{}^{\sourceind{0}}$ true vs $\sourceind{1}$ via $\calest{1}$ using 3 Sources}
        \label{subfig iProNC_Analytic_HFPredEstCalsLF1All}
    \end{subfigure}

    
    \begin{subfigure}{0.83\textwidth}
        \centering
        \includegraphics[width=1.0\linewidth, trim=6 10 7 5, clip]{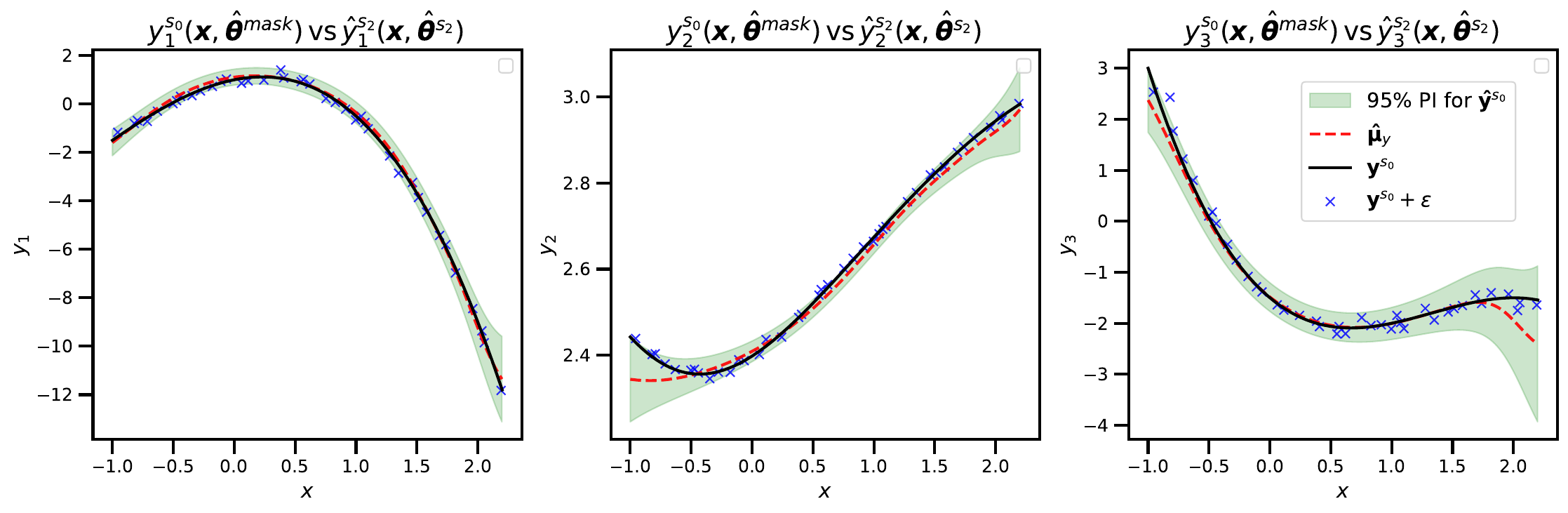}
        \caption{$\indout{}^{\sourceind{0}}$ true vs $\sourceind{2}$ via $\calest{2}$ using 2 Sources}
        \label{subfig iProNC_Analytic_HFPredEstCalsLF2Only}
    \end{subfigure}%
    \hfill
    \begin{subfigure}{0.83\textwidth}
        \centering
        \includegraphics[width=1.0\linewidth]{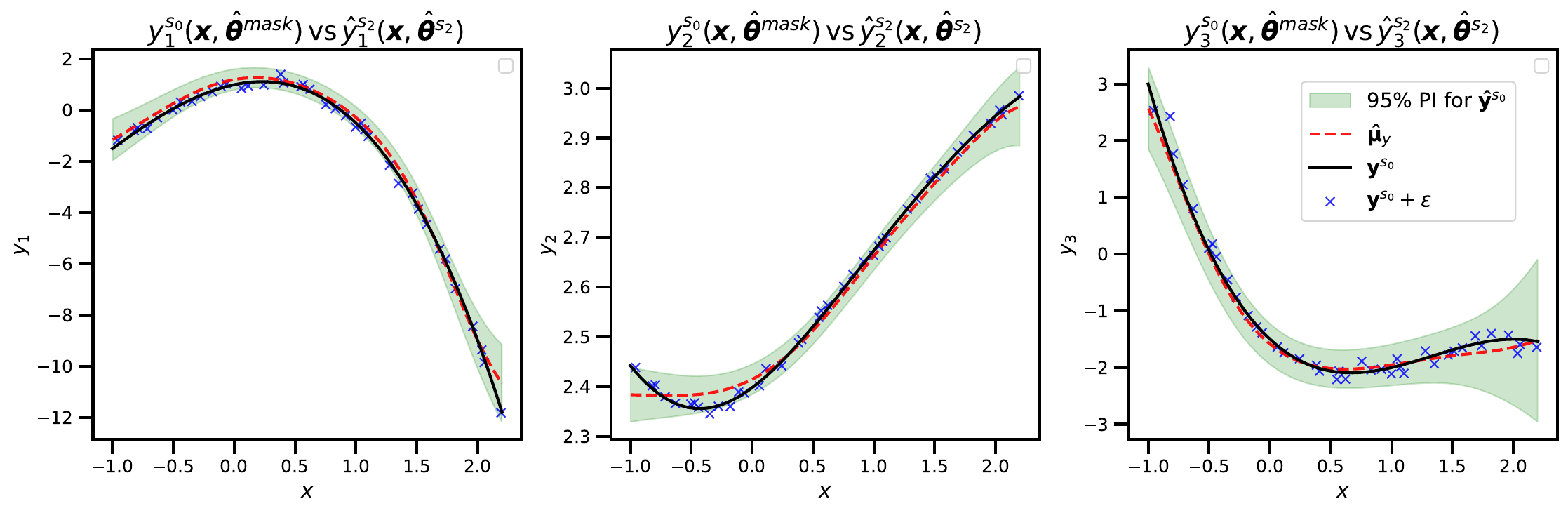}
        \caption{$\indout{}^{\sourceind{0}}$ true vs $\sourceind{2}$ via $\calest{2}$ using 3 Sources}
        \label{subfig iProNC_Analytic_HFPredEstCalsLF2All}
    \end{subfigure}
    \vspace{-0.1cm}
    \caption{\textbf{Calibration and bias correction:} LF sources with estimated calibration parameters emulate the HF source quite well.}
    \label{fig iProNC_Analytic_HFPredEstCals}
\end{figure*}
\begin{figure*}[!ht]
    \centering
    \vspace{-0.5cm}
    \begin{minipage}[t]{0.315\textwidth}
        \centering
        \includegraphics[width=1.0\linewidth]{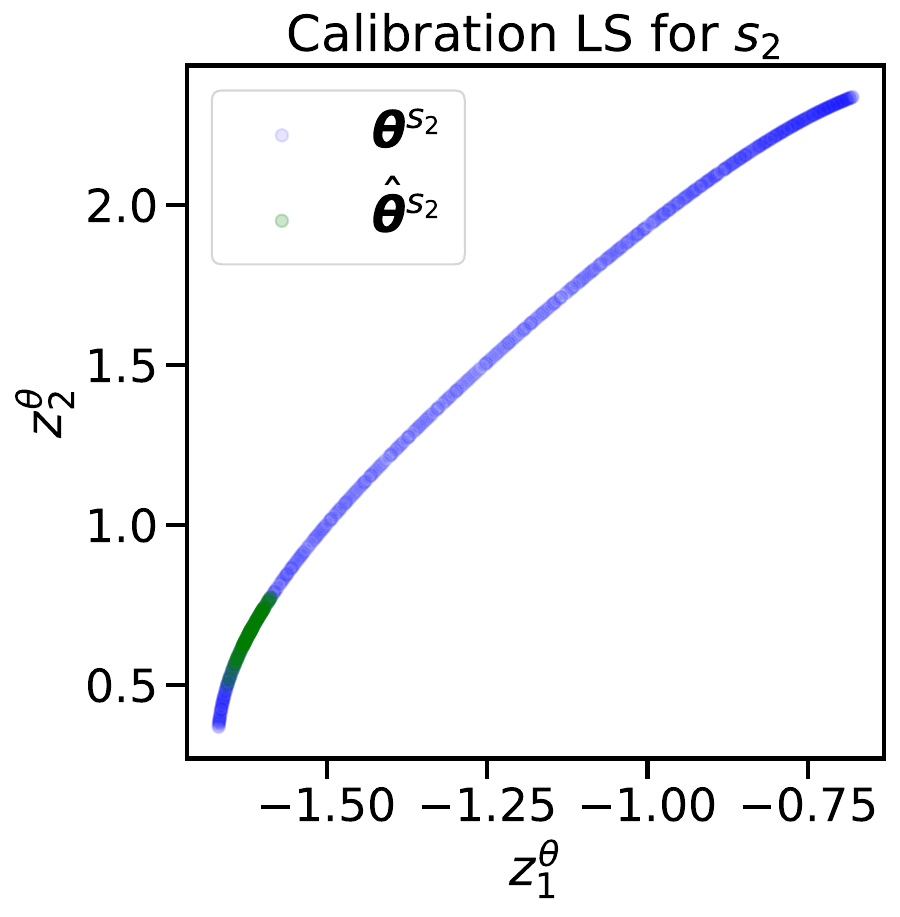}
        \vspace{-0.6cm}
        \subcaption{}
        \label{subfig iProNC_Analytic_CalLSLF2Only}
    \end{minipage}%
    \begin{minipage}[t]{0.32\textwidth}
        \centering
        \includegraphics[width=1.0\linewidth]{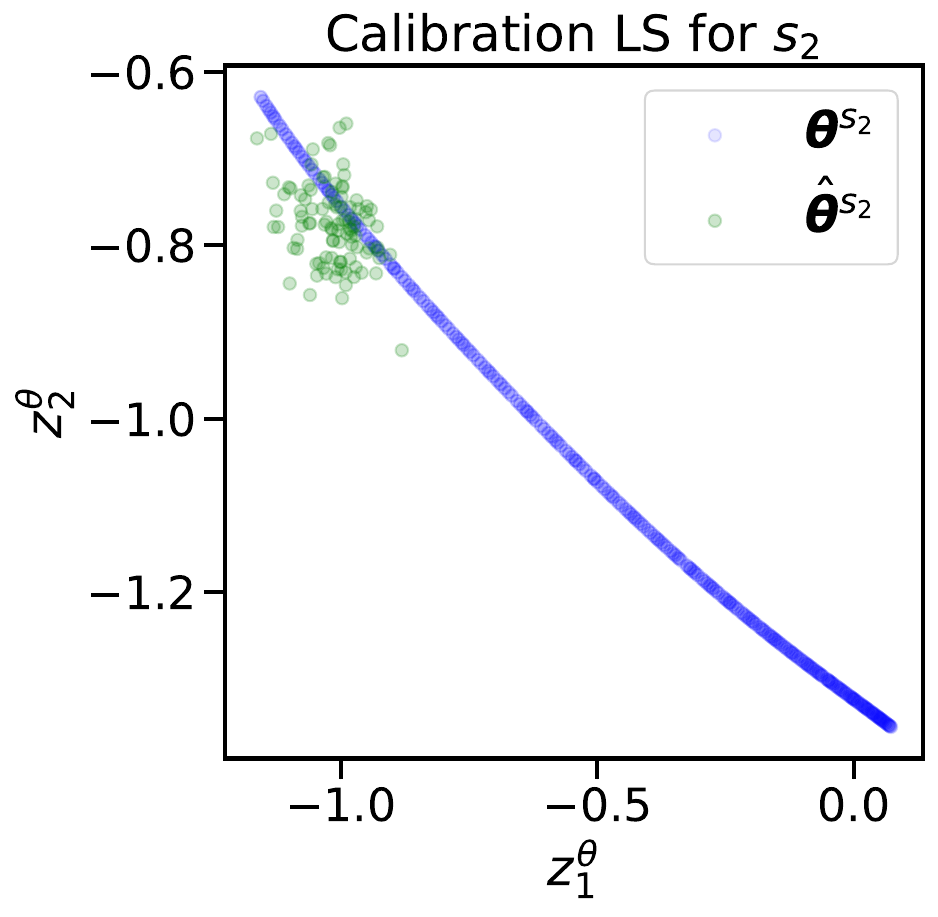}
        \vspace{-0.6cm}
        \subcaption{}
        \label{subfig iProNC_Analytic_CalLSLF2All}
    \end{minipage}%
    \begin{minipage}[t]{.315\textwidth}
        \centering
        \includegraphics[width=1.0\linewidth]{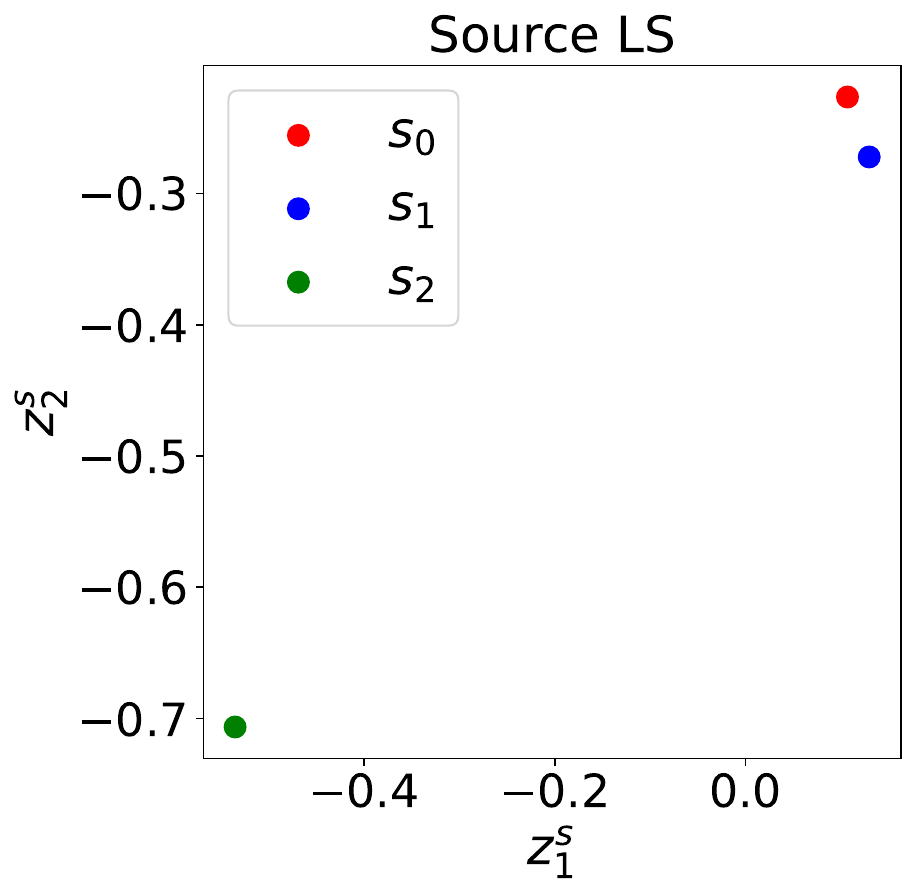}
        \vspace{-0.7cm}
        \subcaption{}
        \label{subfig iProNC_Analytic_SourceLSAll}
    \end{minipage}%
    \vspace{-0.1cm}
    \caption{\textbf{Learned latent spaces:} (a) Block 1 output for $\sourceind{2}$ using 2 Sources. (b) Block 1 output for $\sourceind{2}$ using 3 Sources (c) Block 0 output.}
    \label{fig iProNC_Analytic_CalLSandSourceLS}
\end{figure*}

We next study the performance of \name{} in estimating the calibration parameters. The priors and obtained posterior distributions are shown in Figure~\ref{fig iProNC_Analytic_CalDist} which indicates that the posteriors substantially differ from the priors and cover $\thetabhatmse$ which are the values that minimize the MSE-based discrepancy between the LF sources and the HF source. 
Specifically, we observe in Figure \ref{subfig iProNC_Analytic_CalDistLF1Only} that when $\sourceind{1}$ is calibrated alone, the posterior modes match with $\thetabhatmse$ for only one of the parameters as $\sourceind{1}$ has model-form error. However, we see in Figure \ref{subfig iProNC_Analytic_CalDistLF2Only} that when the network is trained on sources $\sourceind{0}$ and $\sourceind{2}$, it can very closely match the distribution of $\thetashatind{2}{1}$ to $\thetabhatmse$ which is the ground truth in this case as $\sourceind{2}$ does not have any model-form error. Compared to this latter case, training \name{} on all sources reduces the accuracy (see Figure \ref{subfig iProNC_Analytic_CalDistAll}) which is due to the fact that $\sourceind{1}$ has model-form error and its inclusion in the process further complicates calibration.

To further assess the performance of \name{} in calibration, we plot its predictions for each response when an LF source is calibrated either alone or along with the other LF source. Figure \ref{fig iProNC_Analytic_HFPredEstCals} shows that \name{} is effective in both calibration and bias correction: when either of the LF sources is used to emulate the HF one, the predictions of \name{} match with the HF source quite well.

To further show the power of \name{} in bias correction, in Figures \ref{subfig iProNC_Analytic_HFPredEstCalsLF1Only} and \ref{subfig iProNC_Analytic_HFPredEstCalsLF1All} we plot the responses of $\sourceind{1}$ (which has model-form error) by setting the calibration parameters in Equation~\ref{eq 11.1}, ~\ref{eq 11.2}, and ~\ref{eq 11.3} to the estimated values. Comparing the solid magenta, dashed red, and solid black curves in these figures illustrates the power of \name{} in bias correction. 

Finally, we analyze the latent spaces learned by \name{} to assess their interpretability. Sample learned latent spaces are shown in Figure \ref{fig iProNC_Analytic_CalLSandSourceLS} which visualize the similarity of the data sources and the effect of calibration parameters on it. Specifically, Figure \ref{subfig iProNC_Analytic_SourceLSAll} shows the output of Block 0 when \name{} jointly calibrates $\sourceind{1}$ and $\sourceind{2}$. Points in Figure \ref{subfig iProNC_Analytic_SourceLSAll} encode data sources whose similarity is encoded by the distances between the points. We observe that in this particular case \name{} has incorrectly identified $\sourceind{1}$ to be more similar to $\sourceind{0}$ while in reality $\sourceind{2}$ should have been encoded much closer to $\sourceind{0}$ as it does not have model-form error. We attribute this error to the fact that \name{} has very large learning capacity and hence, as seen in Figure \ref{fig iProNC_Analytic_HFPredEstCals}, can correct for model-form errors (via Block 3 and various loss terms). This well-known issue in the literature is commonly referred to as non-identifiability.


In Figures \ref{subfig iProNC_Analytic_CalLSLF2Only} and \ref{subfig iProNC_Analytic_CalLSLF2All} we visualize the encoding that \name{} learns for $\sourceind{2}$ as a function of its calibration parameter. We observe that, expectedly, the encoding based on the posterior distribution covers a smaller region compared to that based on the prior. We also observe the effect of using 2 vs 3 data sources during calibration in these plots where more stochasticity is observed in the posterior encoding corresponding to the latter case. This observation further highlights that calibrating multiple LF sources renders the outcomes of \name{} more uncertain.

\subsection{Engineering Problem} \label{subsec engineering_prob}
\begin{figure}[!h]
    \centering
    \includegraphics[width=0.63\textwidth, trim=0 200 0 200, clip]{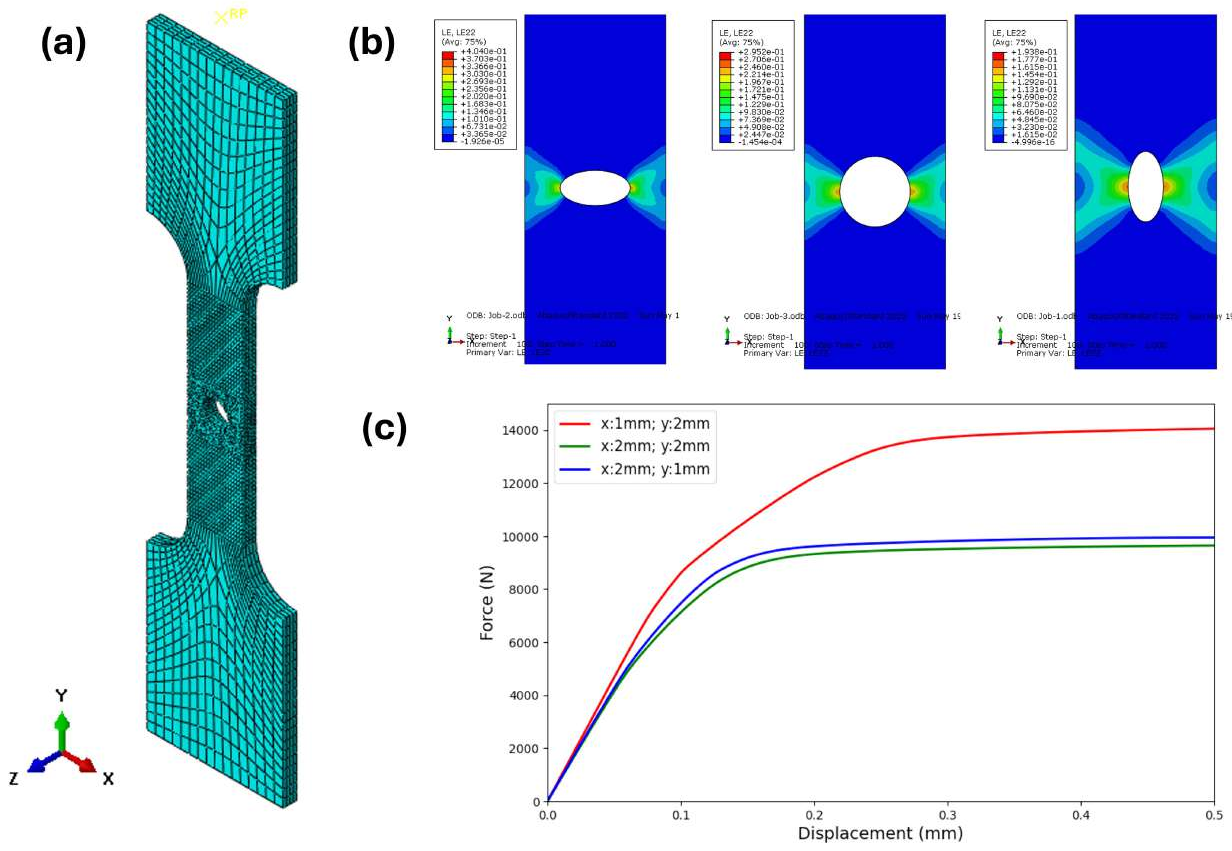}
    \caption{\textbf{Engineering Problem: (a)} A tensile bar with an elliptical hole. 3-D DIC data is obtained in the region around the hole. \textbf{(b)} Strain fields for different elliptical geometries.\textbf{(c)} Global force and displacement curves for tensile bar.}
    \label{fig iProNC_eng_app}
\end{figure}

We evaluate \name{} on a mechanics problem related to material model calibration, see Figure~\ref{fig iProNC_eng_app}. Specifically, we consider a tensile specimen with an elliptical hole whose size varies from one sample to another. For a variety of hole sizes, we simulate the tension test using the Holloman hardening law and two variations of the Voce hardening law \cite{asgharzadeh_determination_2020} for a total of three sources. 
In the former case, we fix the hardening law parameters and treat the resulting simulations as the HF data. For the latter case, we vary the hardening law parameters across the samples and treat the resulting two datasets as LF data. The goal is to calibrate the parameters of the Voce laws such that the resulting FE simulations match those obtained via the Holloman law.

The elastic response of all three material models depends on Young's modulus $E$ and Poisson's ratio $\nu$. The differences are in hardening laws. The Holloman model is defined as:
\begin{equation}
    \sigma_y = \sigma_0 + K(\varepsilon_p)^n
    \label{eq ProNDFCal_holloman}
\end{equation}
where $\sigma_y$ is the yield stress, $\sigma_0$ is the initial yield stress, $\varepsilon_p$ is the equivalent plastic strain, and $K$ and $n$ are material plasticity parameters. The Voce law is given as:
\begin{equation}
    \sigma_y = \sigma_0 + R_0 \varepsilon_p + \sum_{i=1}^{n} R_{\infty,i}(1-e^{-b_i\varepsilon_p})
    \label{eq ProNDFCal_voce}
\end{equation}
where $\sigma_y$, $\sigma_0$, and $\varepsilon_p$
\footnote{We fix $\varepsilon_p$ in both the Holloman and Voce models.} 
are as before, and the summation represents various forms of hardening which can be defined based on the hardening saturation values $R_{\infty,i}$ and hardening rate parameters $b_i$.  
We obtain distinct versions of the Voce model by setting $n=1$ for $\sourceind{1}$ and $n=2$ for $\sourceind{2}$. 
Table~\ref{Table holloman and voce} summarizes the parameters of each hardening law which are either fixed (for Holloman) or have a range that is used for sampling and calibration. 
\begin{table}[!h]
\centering
\caption{\textbf{Hardening law parameters:} Holloman's parameters are fixed for the HF source.}
\label{Table holloman and voce}
\small
\begin{tabular}{@{}ccccc@{}}
\toprule
\multicolumn{5}{c}{\textbf{Holloman Hardening Law Parameters}} \\
\midrule
\textbf{$E$ (\si{GPa})} & \textbf{$\nu$} & \textbf{$\sigma_0$ (\si{MPa})} & \textbf{$K$ (\si{MPa})} & \textbf{$n$} \\
\midrule
206 & 0.26 & 650 & 1500 & 0.36 \\
\bottomrule
\end{tabular}

\vspace{0.5cm} 

\begin{tabular}{@{}cccccc@{}}
\toprule
\multicolumn{6}{c}{\textbf{Voce Hardening Law Parameter Ranges ($i=1, 2$)}} \\
\midrule
\textbf{$E$ (\si{GPa})} & \textbf{$\nu$} & \textbf{$\sigma_0$ (\si{MPa})} & \textbf{$R_{0}$ (\si{MPa})} & \textbf{$R_{\infty, i}$  (\si{MPa})} & \textbf{$b_i$}\\
\midrule
100 - 300 & 0.2 - 0.4 & 400 - 900 & 1000 - 5000 & 0 - 800 & 5 - 500 \\
\bottomrule
\end{tabular}
\end{table}

In addition to the categorical source indicator variable, \name{} takes as inputs (i.e., $\xb$) the major and minor axes of the elliptical hole, displacement of a reference point attached to the top of the bar, and the initial XY coordinates of 12 points chosen on the top left quadrant of the sample's surface. 
The responses (i.e., $\fullout{} = \brackets{y_1, y_2, y_3, y_4}$) include the 3D nodal displacements of the 12 surface nodes and the resulting force at the reference point. We note that the 12 points are chosen to characterize the displacement field and they only span a quarter of the surface due to symmetry. 

To generate data, we apply design of experiments (DoE) to the elliptical hole parameters ($1-2mm$) for all three sources. For the two LF sources, the DoE also includes the calibration parameters listed in Table \ref{Table holloman and voce}. Following this process, we select $n^{s_0}=2000, \ n^{s_1}=8000,$ and $n^{s_2}=8000$ samples from each source. 
We highlight that each bar in the data has a unique mesh, so its nodes do not exactly match the initial 12 XY coordinates we have chosen. To mitigate this issue, we use interpolation which introduces negligible errors into the calibration process.

Since the response of the material to the applied load is extremely different in the elastic and plastic regions, we split the calibration process into two steps. 
We consider three similar scenarios as in Section~\ref{subsec analytic_ex} and use the data from deformation in the elastic region and the beginning of the plastic region for step one and the entire deformation curve in step two.
The first step involves calibrating $E$, $\nu$, and $\sigma_0$ which govern the elastic behavior and its limit. In the second step, we fix the parameters calibrated in step one (except for Poisson ratio) and estimate the remaining calibration parameters in Equation \ref{eq ProNDFCal_voce} (we treated Poisson ratio differently as \name{} provided inconsistent estimates for it across different runs).

Table~\ref{tab eng ex cal uq} shows the performance of \name{} in estimating the calibration parameters using all data sources. We observe that \name{} provides reasonable accuracy for calibrating $E$ (see $\muhatnos{1}$ and $\sigmahatnos{1}$ in columns 1 and 2) and $\sigma_0$ (see columns 5 and 6) but we noticed that these estimates vary depending on the network initialization.

We also observe that \name{} fails to accurately estimate $\nu$ which is somewhat expected because $\nu$ mainly affects out-of-plane (i.e., $z$) displacements in a tension test. 
These displacement are much smaller than the XY displacements hence learning them is more difficult. 
\begin{table}[h]
    \centering
    \caption{\textbf{Calibration results (Step I):} All sources are used for calibrating $E$, $\nu$, and $\sigma_0$ for $\sourceind{1}$ and $\sourceind{2}$.}
    \label{tab eng ex cal uq}
    \small
    \begin{tabular}{@{}c c c c c c c@{}}
        \toprule
        & $\muhatnos{1}$ & $\sigmahatnos{1}$ & $\muhatnos{2}$ & $\sigmahatnos{2}$ & $\muhatnos{3}$ & $\sigmahatnos{3}$\\
        \midrule
        $\thetabshat{1}$ & \num{2.10e+05} & \num{8.40e+03} & 0.397 & 0.011 & 644 & 33.2 \\
        $\thetabshat{2}$ & \num{2.12e+05} & \num{9.02e+03} & 0.397 & 0.011 & 767 & 38.0 \\
        $\thetabhatmse$ & \num{2.06e5} & {--} & 0.260 & {--} & 650 & {--} \\
        \bottomrule
    \end{tabular}
\end{table}

Although we introduce another calibration step to simplify the problem for \name{}, the estimation of the calibration parameters in step two remains highly stochastic. Similarly to the estimation of $\muhatnos{2}$, most of the Voce hardening parameters are estimated to be at the extrema of their sampling ranges. In each of the training cases, there are one or two hardening parameters that seem to approach reasonable distributions, but this is inconsistent and there is no ground truth to compare the results. This stochasticity, observed in both the estimated calibration parameters and the latent spaces, suggests that the higher-dimensional problem suffers from non-identifiability. 
We hypothesize that the calibration performance of the model was adversely impacted by the higher dimensionality of $\thetab$ and $\xb$, and the network's struggle to learn all tasks simultaneously.

\begin{table*}[!h]
    \centering
    \vspace{-0.3cm}
    \caption{\textbf{Errors after Step I Calibration:} RRMSE of \(\yhb\) vs \(\yhbhat\), \(\ybshat{1}\), and \(\ybshat{2}\)}
    \label{tab eng ex first step}
    \small
    \renewcommand{\arraystretch}{1.2} 
    \setlength{\tabcolsep}{4pt} 
    \begin{tabularx}{\textwidth}{|l|*{4}{>{\centering\arraybackslash}X|}}
        \hline
        \textbf{Dataset} 
        & $\yhout{1}$ vs $\yhhatthetagenshat{1}$
        & $\yhout{2}$ vs $\yhhatthetagenshat{2}$
        & $\yhout{3}$ vs $\yhhatthetagenshat{3}$
        & $\yhout{4}$ vs $\yhhatthetagenshat{4}$ \\
        \hline
        $\sourceind{0}$ and $\sourceind{1}$  & 0.13282 & 0.11098 & 0.04523 & 0.26717 \\ \hline
        $\sourceind{0}$ and $\sourceind{1}$  & 0.13656 & 0.17898 & 0.05260 & 0.28243 \\ \hline
        All Sources & 0.03568 & 0.05219 & 0.02041 & 0.06388 \\ \hline
        \textbf{Dataset} 
        & $\yhout{1}$ vs $\yshatthetashat{1}{1}$
        & $\yhout{2}$ vs $\yshatthetashat{2}{1}$
        & $\yhout{3}$ vs $\yshatthetashat{3}{1}$
        & $\yhout{4}$ vs $\yshatthetashat{4}{1}$ \\ 
        \hline
        $\sourceind{0}$ and $\sourceind{1}$ & 1.79507 & 1.50945 & 1.71485 & 0.06393 \\ \hline
        All Sources & 0.02749 & 0.04849 & 0.01992 & 0.07023  \\ \hline
        \textbf{Dataset} 
        & $\yhout{1}$ vs $\yshatthetashat{1}{2}$
        & $\yhout{2}$ vs $\yshatthetashat{2}{2}$
        & $\yhout{3}$ vs $\yshatthetashat{3}{2}$
        & $\yhout{4}$ vs $\yshatthetashat{4}{2}$ \\ 
        \hline
        $\sourceind{0}$ and $\sourceind{2}$ & 1.64083 & 1.47877 & 1.63542 & 0.07674  \\ \hline
        All Sources & 0.02868 & 0.04788 & 0.02006 & 0.07056  \\ \hline
    \end{tabularx}
\end{table*}

\name{} performs quite well in terms of emulation even though it fails to consistently estimate the material properties accurately. We attribute this feature to the networks ability to do bias correction and show it in Tables \ref{tab eng ex first step} and \ref{tab eng ex second step} which summarize the emulation accuracy after each step of our two-step calibration approach. 
Comparing rows 2-4 in Table~\ref{tab eng ex first step}, we see that the inclusion of all three sources significantly improves LF emulation of the HF source. Similarly, comparing rows 6-7 (for $\sourceind{1}$) or rows 9-10 (for $\sourceind{2}$) we observe that an LF surrogate with estimated calibration parameters can emulate the HF source more accurately when all the data are used for calibration.
From the results in Table~\ref{tab eng ex second step}, we see that emulating the HF source by setting $t_s = s_0$ (rows 2-4) is significantly better than emulating the HF source via calibrated LFs (rows 6-7 for $t_s = s_1$ and 9-10 for $t_s = s_2$). 

\begin{table*}[!h]
    \centering
    \caption{\textbf{Errors after Step II Calibration:} RRMSE of \(\yhb\) vs \(\yhbhat\), \(\ybshat{1}\), and \(\ybshat{2}\)}
    \label{tab eng ex second step}
    \renewcommand{\arraystretch}{1.2} 
    \setlength{\tabcolsep}{4pt} 
    \small
    \begin{tabularx}{\textwidth}{|l|*{4}{>{\centering\arraybackslash}X|}}
        \hline
        \textbf{Dataset} 
        & $\yhout{1}$ vs $\yhhatthetagenshat{1}$
        & $\yhout{2}$ vs $\yhhatthetagenshat{2}$
        & $\yhout{3}$ vs $\yhhatthetagenshat{3}$
        & $\yhout{4}$ vs $\yhhatthetagenshat{4}$ \\
        \hline
        $\sourceind{0}$ and $\sourceind{1}$  & 0.38802 & 0.62677 & 0.64960 & 0.79586 \\ \hline
        $\sourceind{0}$ and $\sourceind{2}$  & 0.12018 & 0.11422 & 0.04938 & 0.11286 \\ \hline
        All Sources & 0.24249 & 0.11590 & 0.07363 & 0.19953 \\ \hline
        \textbf{Dataset} 
        & $\yhout{1}$ vs $\yshatthetashat{1}{1}$
        & $\yhout{2}$ vs $\yshatthetashat{2}{1}$
        & $\yhout{3}$ vs $\yshatthetashat{3}{1}$
        & $\yhout{4}$ vs $\yshatthetashat{4}{1}$ \\ 
        \hline
        $\sourceind{0}$ and $\sourceind{1}$ & 1.35483 & 6.37452 & 4.01805 & 10.22809 \\ \hline
        All Sources & 1.87659 & 1.40288 & 1.57942 & 0.02995  \\ \hline
        \textbf{Dataset} 
        & $\yhout{1}$ vs $\yshatthetashat{1}{2}$
        & $\yhout{2}$ vs $\yshatthetashat{2}{2}$
        & $\yhout{3}$ vs $\yshatthetashat{3}{2}$
        & $\yhout{4}$ vs $\yshatthetashat{4}{2}$ \\ 
        \hline
        $\sourceind{0}$ and $\sourceind{2}$ & 1.96485 & 1.71372 & 1.79920 & 0.05350  \\ \hline
        All Sources & 1.89350 & 1.41155 & 1.59264 & 0.03403  \\ \hline
    \end{tabularx}
\end{table*}








    \section{Discussion} \label{sec discussion}

In developing \name{}, we have explored a number of schemes to represent and estimate the calibration parameters, as well as a number of variations in the architecture presented in Section~\ref{sec method}. We believe it is useful to discuss some of our efforts and observations below.

It was essential to develop a strategy to handle missing calibration parameters.
One method that we diverged from involved replacing the estimated parameters of the HF source by random data. In this approach, we introduced a term to the loss function that was the Jacobian of Block 1 outputs with respect to the HF calibration inputs, \ie we encouraged Block 1 to learn a mapping that is independent of the calibration input. However, this approach requires Block 1 to learn two entirely disparate tasks: $(1)$ to be highly sensitive to the calibration input for LF sources, and $(2)$ be insensitive to the calibration input for the HF source. This disparity compromised the overall accuracy of the network and forced Block 0 to place the HF and LF sources distant from each other even if there was no model form error. The Jacobian loss term also introduced an additional hyperparameter which required tuning and increased computational costs.

We have also experimented with techniques to learn the distributions of the calibration parameters via an NN block. We first tried generating artificial random data to serve as calibration inputs and feeding them through an NN block. The output of this NN would represent the calibration estimates and are fed to Block 1. The posterior distribution for the calibration estimates could then be obtained by feeding data drawn from this same random distribution through the block. This approach increased the size of the network, adding additional parameters and increasing the risk of over-fitting. It also required including a separate approach for handling missing calibration parameters such as the Jacobian term mentioned above.

Due to the large number of tasks that \name{} must learn, the model is very complex and its performance depends on initialization. We observed that adjusting the calibration process based on the physics of the problem, as done in Section \ref{subsec engineering_prob}, substantially improves the performance of the model. Regarding the multi-task nature of the loss, we tried automatic loss weighting but this approach did not consistently improve the performance across various tests and as a result we excluded it from the final model configuration. 
Further analysis of this behavior is important especially because we observed that during training the network learns to emulate the HF source (when setting $t_s=s_0$) faster than all other emulation and calibration tasks. Some tasks might not be learned at all and this leads to higher computational costs and lower accuracy.



Throughout the development of \name{}, we experimented with the size of each block. We found the best performance when the size of Blocks 0, 1, and 2 is set to have one layer containing five neurons. We also observed that the size of Block 3 and the batch size of the training data affect the performance. We obtained the best performance when Block 3 had its layers expand towards the center of the architecture and contract towards the output layer. 

Finally, we stress the need to examine the computational cost of calibrating more than two sources at once. The cost of training \name{} on the \ExIIname~ using two sources of data takes an average of 21 minutes. Training three sources takes approximately 31 minutes. There is only a marginal decrease in calibration performance using more than two sources of data on low dimensional problems. For these types of problems, it would be more advantageous to calibrate an arbitrary number of sources at once especially given the performance boost in emulation. 


    \section{Conclusion} \label{sec conclusion}

We introduce \name{} to simultaneously emulate and calibrate any number of computer models. \name{} is built on a customized multi-block NN architecture that learns interpretable information about multi-response models and their tuning parameters as well as fidelity levels. Our method learns probabilistic distributions for system responses and calibration parameters, providing separable measures of aleatoric and epistemic uncertainty, respectively. The probability distributions of each calibration parameter are independent and unique to each model which enables \name{} to not only identify latent relationships in the data, but also accommodate applications where computer models have different number of calibration parameters.

Our study shows that \name{} has the potential to be a powerful data fusion approach that can uncover hidden correlations and behaviors in a variety of different systems. We also evaluate its performance in situations where no prior knowledge or biases are given and conclude that domain knowledge must be included into the model in high-dimensional and complex applications to avoid overfitting and non-identifiability issues. Due to the high cost of architectural tuning, our conclusion is that using \name{} is only justified if maximizing emulation accuracy is the only goal. 
    \section*{Acknowledgments}
We appreciate the support from Office of Naval Research (grant number $N000142312485$), the National Science Foundation (grant numbers $2238038$ and $2525731$), and NASA’s Space Technology Research Grants Program (grant number $80NSSC21K1809$).
    \clearpage
\onecolumn
\appendix
\section{Notation Guide} \label{subsec notation}

Bold capital letters, like  $\fullout{}$, are considered matrices. Bold lowercase letters are considered vectors, like $\numinp{}$. Letters with an unmodified font are scalars. The superscript $^{s_{i}}$ is used to denote the $i^{th}$ data source. We may also use $j$ to index the data sources if $i$ has been reserved for another purpose. For instance, $i$ will be used to denote the specific outputs of a system, as in $\indout{i}$. The superscript $^s$, in $\zb^s$, and the subscript $_s$, in  $t_s$, are used to denote variables referring to the data sources. Similarly, the superscript $^c$, in $\zb^c$, and the subscript $_c$, in  $t_c$, are used to denote variables referring to the categorical variables. Superscripts enclosed by parentheses, as in $^{(k)}$, denote the $k^{th}$ sample.

\section*{Nomenclature}
\begin{description}
\item[$\eta^{h}{}(\cdot)$] GP Emulator of HF Source
\item[$\eta^{l}(\cdot)$] GP Emulator of LF Source
\item[$\delta(\cdot)$] GP Emulator of Discrepancy/Bias Function
\item[$\thetab{}^*$] True Calibration Parameters
\item[$N$] Number of Samples (e.g. in a training batch)
\item[$n_y$] Number of Outputs
\item[$n^{s_i}$] Number of Samples in the $i^{th}$ Source
\item[$\boldsymbol{\sigma}^{2,~s_0}$] HF Noise

\item[$\numinp{}$] System Inputs
\item[$\zeta(\cdot)$] Deterministic Encoder Function
\item[$t_s$] Source Indicator Variable
\item[$\tb_c$] Categorical Variable
\item[$\thetab{}$] Calibration Parameters
\item[$\hat{\boldsymbol{\phi}}$] Additional Model Inputs (e.g. $\OHenc{\sourceinp{} = \sourceind{j}}, \catinps{j}$) and Model Parameters
\item[$ds$] Number of Data Sources

\item[$\fullout{}$] Output Matrix
\item[$\indout{i}$] $i^{th}$ Output of the Entire Dataset
\item[$\ysbindoutk{j}{i}$] $k^{th}$ sample of the $i^{th}$ Output of the $j^{th}$ Source

\item[$\indoutmean{i}$] Estimated Mean of $i^{th}$ Output
\item[$\indoutstd{i}$] Estimated Standard Deviation of $i^{th}$ Output
\item[$\calest{i}$] Estimated Calibration Parameters for Source $s_i$]

\item[$\zb^s$] Source Latent Variables
\item[$\zcal{}$] Calibration Parameter Latent Variables
\item[$\zcat{}$] Categorical Latent Variables
\newline 
\end{description}

    \pagebreak
    \bibliography{R_Ref}

\begin{thebibliography}{10}

\bibitem{sacks_design_1989}
Jerome Sacks, William~J. Welch, Toby~J. Mitchell, and Henry~P. Wynn.
\newblock Design and {Analysis} of {Computer} {Experiments}.
\newblock {\em Statistical Science}, 4(4):409--423, November 1989.
\newblock Publisher: Institute of Mathematical Statistics.

\bibitem{forest_inferring_2008}
Chris~E. Forest, Bruno Sansó, and Daniel Zantedeschi.
\newblock Inferring climate system properties using a computer model.
\newblock {\em Bayesian Analysis}, 3(1):1--37, March 2008.
\newblock Publisher: International Society for Bayesian Analysis.

\bibitem{salter_uncertainty_2019}
James~M. Salter, Daniel~B. Williamson, John Scinocca, and Viatcheslav Kharin.
\newblock Uncertainty {Quantification} for {Computer} {Models} {With} {Spatial} {Output} {Using} {Calibration}-{Optimal} {Bases}, October 2019.
\newblock Publisher: Taylor \& Francis.

\bibitem{larssen_forecasting_2006}
Thorjørn Larssen, Ragnar~B. Huseby, Bernard~J. Cosby, Gudmund Høst, Tore Høgåsen, and Magne Aldrin.
\newblock Forecasting {Acidification} {Effects} {Using} a {Bayesian} {Calibration} and {Uncertainty} {Propagation} {Approach}.
\newblock {\em Environmental Science \& Technology}, 40(24):7841--7847, December 2006.

\bibitem{arhonditsis_eutrophication_2007}
George~B. Arhonditsis, Song~S. Qian, Craig~A. Stow, E.~Conrad Lamon, and Kenneth~H. Reckhow.
\newblock Eutrophication risk assessment using {Bayesian} calibration of process-based models: {Application} to a mesotrophic lake.
\newblock {\em Ecological Modelling}, 208(2-4):215--229, November 2007.

\bibitem{henderson_bayesian_2009}
Daniel~A. Henderson, Richard~J. Boys, Kim~J. Krishnan, Conor Lawless, and Darren~J. Wilkinson.
\newblock Bayesian {Emulation} and {Calibration} of a {Stochastic} {Computer} {Model} of {Mitochondrial} {DNA} {Deletions} in {Substantia} {Nigra} {Neurons}.
\newblock {\em Journal of the American Statistical Association}, 104(485):76--87, March 2009.

\bibitem{gattiker_combining_2006}
Jim Gattiker, Dave Higdon, Sallie Keller-McNulty, Michael McKay, Leslie Moore, and Brian Williams.
\newblock Combining experimental data and computer simulations, with an application to flyer plate experiments.
\newblock {\em Bayesian Analysis}, 1(4):765--792, December 2006.
\newblock Publisher: International Society for Bayesian Analysis.

\bibitem{baltic_2021_MLFractureLocusCalibration}
Sandra Baltic, Mohammad~Zhian Asadzadeh, Patrick Hammer, Julien Magnien, Hans-Peter Gänser, Thomas Antretter, and René Hammer.
\newblock Machine learning assisted calibration of a ductile fracture locus model.
\newblock {\em Materials \& Design}, 203:109604, May 2021.

\bibitem{abaqus_documentation}
Michael Smith.
\newblock {\em ABAQUS/Standard User's Manual, Version 6.9}.
\newblock Dassault Syst{\`e}mes Simulia Corp, United States, 2009.

\bibitem{Rasmussen2006Gaussian}
Carl Rasmussen and Christopher Williams.
\newblock {\em Gaussian Processes For Machine Learning}.
\newblock The MIT Press, 2006.

\bibitem{eweis2022data}
Jonathan~Tammer Eweis-Labolle, Nicholas Oune, and Ramin Bostanabad.
\newblock Data fusion with latent map gaussian processes.
\newblock {\em Journal of Mechanical Design}, 144(9):091703, 2022.

\bibitem{deng2023data}
Shiguang Deng, Carlos Mora, Diran Apelian, and Ramin Bostanabad.
\newblock Data-driven calibration of multifidelity multiscale fracture models via latent map gaussian process.
\newblock {\em Journal of Mechanical Design}, 145(1):011705, 2023.

\bibitem{yousefpour_gp_2023}
Amin Yousefpour, Zahra~Zanjani Foumani, Mehdi Shishehbor, Carlos Mora, and Ramin Bostanabad.
\newblock {GP}+: {A} {Python} {Library} for {Kernel}-based learning via {Gaussian} {Processes}, December 2023.
\newblock arXiv:2312.07694 [cs, stat].

\bibitem{planas_extrapolation_2020}
Robert Planas, Nicholas Oune, and Ramin Bostanabad.
\newblock Extrapolation {With} {Gaussian} {Random} {Processes} and {Evolutionary} {Programming}, 2020.

\bibitem{kennedy2001bayesian}
Marc~C Kennedy and Anthony O'Hagan.
\newblock Bayesian calibration of computer models.
\newblock {\em Journal of the Royal Statistical Society: Series B (Statistical Methodology)}, 63(3):425--464, 2001.

\bibitem{higdon2004combining}
Dave Higdon, Marc Kennedy, James~C Cavendish, John~A Cafeo, and Robert~D Ryne.
\newblock Combining field data and computer simulations for calibration and prediction.
\newblock {\em SIAM Journal on Scientific Computing}, 26(2):448--466, 2004.

\bibitem{plumlee2017bayesian}
Matthew Plumlee.
\newblock Bayesian calibration of inexact computer models.
\newblock {\em Journal of the American Statistical Association}, 112(519):1274--1285, 2017.

\bibitem{zhang2019numerical}
Weizhao Zhang, Ramin Bostanabad, Biao Liang, Xuming Su, Danielle Zeng, Miguel~A Bessa, Yanchao Wang, Wei Chen, and Jian Cao.
\newblock A numerical bayesian-calibrated characterization method for multiscale prepreg preforming simulations with tension-shear coupling.
\newblock {\em Composites Science and Technology}, 170:15--24, 2019.

\bibitem{apley2006understanding}
Daniel~W Apley, Jun Liu, and Wei Chen.
\newblock Understanding the effects of model uncertainty in robust design with computer experiments, 2006.

\bibitem{bayarri2007framework}
Maria~J Bayarri, James~O Berger, Rui Paulo, Jerry Sacks, John~A Cafeo, James Cavendish, Chin-Hsu Lin, and Jian Tu.
\newblock A framework for validation of computer models.
\newblock {\em Technometrics}, 49(2):138--154, 2007.

\bibitem{arendt2012quantification}
Paul~D Arendt, Daniel~W Apley, and Wei Chen.
\newblock Quantification of model uncertainty: Calibration, model discrepancy, and identifiability, 2012.

\bibitem{arendt2012improving}
Paul~D Arendt, Daniel~W Apley, Wei Chen, David Lamb, and David Gorsich.
\newblock Improving identifiability in model calibration using multiple responses, 2012.

\bibitem{stainforth2005uncertainty}
David~A Stainforth, Tolu Aina, Carl Christensen, Mat Collins, Nick Faull, Dave~J Frame, Jamie~A Kettleborough, S~Knight, A~Martin, JM~Murphy, et~al.
\newblock Uncertainty in predictions of the climate response to rising levels of greenhouse gases.
\newblock {\em Nature}, 433(7024):403--406, 2005.

\bibitem{gramacy2015calibrating}
Robert~B Gramacy, Derek Bingham, James~Paul Holloway, Michael~J Grosskopf, Carolyn~C Kuranz, Erica Rutter, Matt Trantham, and R~Paul Drake.
\newblock Calibrating a large computer experiment simulating radiative shock hydrodynamics.
\newblock {\em The Annals of Applied Statistics}, 9(3):1141--1168, 2015.

\bibitem{RN648}
Rui Tuo and CF~Wu.
\newblock Prediction based on the kennedy-o'hagan calibration model: asymptotic consistency and other properties, 2017.

\bibitem{RN1071}
Rui Tuo.
\newblock Adjustments to computer models via projected kernel calibration.
\newblock {\em SIAM/ASA Journal on Uncertainty Quantification}, 7(2):553--578, 2019.

\bibitem{RN271}
Paul~D Arendt, Daniel~W Apley, Wei Chen, David Lamb, and David Gorsich.
\newblock Improving identifiability in model calibration using multiple responses.
\newblock {\em Journal of Mechanical Design}, 134(10):100909, 2012.

\bibitem{RN664}
Zhen Jiang, Wei Chen, and Daniel~W Apley.
\newblock Preposterior analysis to select experimental responses for improving identifiability in model uncertainty quantification.
\newblock In {\em ASME 2013 International Design Engineering Technical Conferences and Computers and Information in Engineering Conference}, pages V03BT03A051--V03BT03A051. American Society of Mechanical Engineers, 2013.

\bibitem{higdon_HDO_2008}
Dave Higdon, James Gattiker, Brian Williams, and Maria Rightley.
\newblock Computer model calibration using high dimensional output, 2008.

\bibitem{10.5555/3045390.3045502}
Yarin Gal and Zoubin Ghahramani.
\newblock Dropout as a bayesian approximation: Representing model uncertainty in deep learning.
\newblock In {\em Proceedings of the 33rd International Conference on International Conference on Machine Learning - Volume 48}, ICML'16, page 1050–1059. JMLR.org, 2016.

\bibitem{RN1074}
N.~Srivastava, G.~Hinton, A.~Krizhevsky, I.~Sutskever, and R.~Salakhutdinov.
\newblock Dropout: A simple way to prevent neural networks from overfitting.
\newblock {\em Journal of Machine Learning Research}, 15(1):1929--1958, 2014.

\bibitem{RN1191}
Ehsan Kharazmi, Zhongqiang Zhang, and George~Em Karniadakis.
\newblock Variational physics-informed neural networks for solving partial differential equations, 2019.

\bibitem{blundell2015weight}
Charles Blundell, Julien Cornebise, Koray Kavukcuoglu, and Daan Wierstra.
\newblock Weight uncertainty in neural network.
\newblock In {\em International conference on machine learning}, pages 1613--1622. PMLR, 2015.

\bibitem{lecun2015deep}
Yann LeCun, Yoshua Bengio, and Geoffrey Hinton.
\newblock Deep learning.
\newblock {\em Nature}, 521(7553):436--444, May 2015.
\newblock Publisher: Nature Publishing Group.

\bibitem{hamel_calibrating_2023}
Craig~M. Hamel, Kevin~N. Long, and Sharlotte L.~B. Kramer.
\newblock Calibrating constitutive models with full-field data via physics informed neural networks.
\newblock {\em Strain}, 59(2):e12431, April 2023.
\newblock Publisher: John Wiley \& Sons, Ltd.

\bibitem{yousefpour_GPPINNs_2024}
Amin Yousefpour, Shirin Hosseinmardi, Carlos Mora, and Ramin Bostanabad.
\newblock Simultaneous and meshfree topology optimization with physics-informed gaussian processes, 2024.

\bibitem{mora_eweis2023prondf}
Carlos Mora, Jonathan~Tammer Eweis-Labolle, Tyler Johnson, Likith Gadde, and Ramin Bostanabad.
\newblock Probabilistic neural data fusion for learning from an arbitrary number of multi-fidelity data sets.
\newblock {\em Computer Methods in Applied Mechanics and Engineering}, 415:116207, 2023.

\bibitem{10.5555/3495724.3496975}
Alexander Amini, Wilko Schwarting, Ava Soleimany, and Daniela Rus.
\newblock Deep evidential regression.
\newblock In {\em Proceedings of the 34th International Conference on Neural Information Processing Systems}, NIPS'20, pages 14927 -- 14937, Red Hook, NY, USA, 2020. Curran Associates Inc.

\bibitem{loshchilov_decoupled_2019}
Ilya Loshchilov and Frank Hutter.
\newblock Decoupled {Weight} {Decay} {Regularization}, January 2019.
\newblock arXiv:1711.05101 [cs, math].

\bibitem{RN941}
Diederik~P Kingma and Max Welling.
\newblock Auto-encoding variational bayes, 2013.

\bibitem{PyTorch_Lightning}
William Falcon and The PyTorch~Lightning team.
\newblock Pytorch lightning, March 2019.
\newblock license: "Apache-2.0", repository-code: "https://github.com/Lightning-AI/lightning", version: "1.4".

\bibitem{asgharzadeh_determination_2020}
Amir Asgharzadeh, Sobhan~Alah Nazari~Tiji, Rasoul Esmaeilpour, Taejoon Park, and Farhang Pourboghrat.
\newblock Determination of hardness-strength and -flow behavior relationships in bulged aluminum alloys and verification by {FE} analysis on {Rockwell} hardness test.
\newblock {\em The International Journal of Advanced Manufacturing Technology}, 106(1-2):315--331, January 2020.

\end{thebibliography}
    
    
    
\end{document}